\documentclass{article} 
\usepackage{iclr2027_conference,times}

\usepackage{amsmath,amsfonts,bm}

\def\eqref#1{equation~\ref{#1}}

\def\1{\bm{1}}

\DeclareMathAlphabet{\mathsfit}{\encodingdefault}{\sfdefault}{m}{sl}
\SetMathAlphabet{\mathsfit}{bold}{\encodingdefault}{\sfdefault}{bx}{n}

\usepackage{hyperref}
\usepackage{url}
\usepackage{graphicx}
\usepackage{booktabs}
\usepackage{multirow}
\usepackage{amsmath}
\usepackage{amsfonts}  
\usepackage[most]{tcolorbox}
\usepackage{listings}
\usepackage{tabularx}
\usepackage{stfloats} 
\usepackage{textcomp}
\usepackage{colortbl}
\usepackage{xcolor}
\usepackage{hhline}
\usepackage{amssymb}

\newcommand{\ours}{TCR-Bench}
\title{Semantically Similar, Yet Not Answerable: Diagnosing the Semantic-Answerability Gap in Table RAG}

\author{
Jiaming Tian$^{1}$ \quad
Liyao Li$^{1}$ \quad
Wentao Ye$^{1}$ \quad
Haobo Wang$^{1}$\thanks{Corresponding author.} \quad
Peng Lu$^{1}$ \quad \\
Lihua Yu$^{2}$ \quad
Zujie Ren$^{1,3}$ \quad
Gang Chen$^{1}$ \quad
Junbo Zhao$^{1}$ \\
$^{1}$Zhejiang University \quad
$^{2}$Bank of Hangzhou Co., Ltd. \quad
$^{3}$Zhejiang Lab \\
\texttt{hsxz2@zju.edu.cn, wanghaobo@zju.edu.cn}
}

\iclrfinalcopy 
\begin{document}

\maketitle
\begin{abstract}
In retrieval-augmented generation (RAG), \textbf{semantic relevance} asks whether a source matches a query in meaning, while \textbf{answerability} asks whether it contains sufficient information to answer the query. A \textbf{Semantic-Answerability Gap (SAG)} may arise in retrieval when a retriever can reach semantically relevant sources yet fail to identify those that are uniquely answerable. We uncover this gap using tables as a controlled setting, where shared schemas and entities provide strong semantic signals while localized content and row-column bindings distinguish answerable from non-answerable sources. Using \textbf{\ours{}}, a controlled sibling-table benchmark, we find that dense retrievers achieve only \textbf{18.2\%} top-1 target retrieval, reducing QA F1 from \textbf{0.755} with the oracle table to \textbf{0.330} with retrieved top-5 tables. Controlled diagnostics show that retrievers favor semantic volume over sufficiency, respond weakly to row-column binding disruptions, and struggle to distinguish Targets from Siblings. Explicit answerability assessment substantially improves target identification, while fine-tuning shows that answerability is learnable but difficult to transfer without compromising broad semantic retrieval.
\end{abstract}

\section{Introduction}

In retrieval-augmented generation (RAG) systems~\citep{firstRAG,RAG-survey}, retrieval must do more than find sources relevant to a query: the retrieved source should contain sufficient information to support a valid answer. We refer to this property as \textbf{answerability}, which is distinct from semantic relevance, or whether a source matches the meaning or topic of a query. A source may therefore be relevant without being sufficient to answer the query. This raises a natural question: can retrieval identify sources that are both relevant and answerable?

\begin{figure}[b]
\centering
\includegraphics[width=0.95\linewidth]{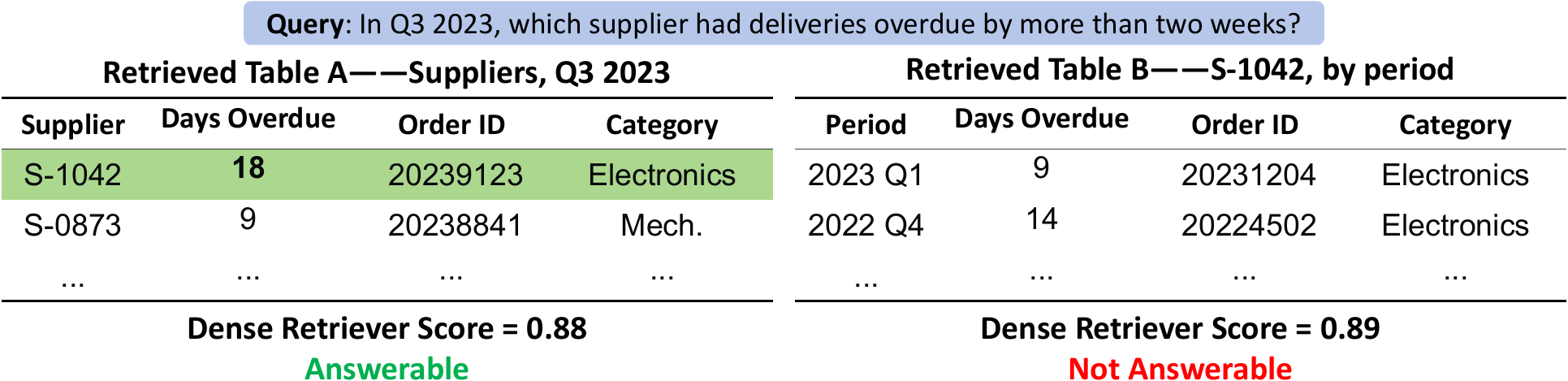}
\caption{Two sibling tables with nearly identical schemas yield similar retrieval signals but differ in answerability: Table A contains the answer, while Table B scores higher because its value (14 days) is semantically close to the query's threshold (two weeks), despite belonging to the wrong period.}
\label{fig:answerability_example}
\end{figure}

We study this question using tables as a controlled setting. Their structured, localized information allows us to construct sources that are highly similar in semantics and structure but differ in whether they contain the information needed to answer a query. This makes tables a useful setting for disentangling semantic relevance from answerability. Moreover, tables are widely used as information sources in enterprise databases, data lakes, and web resources~\citep{table-in-web,table-in-system,table-in-lake}, where semantically similar tables may likewise differ in the information they contain.

Existing table QA benchmarks such as WikiTableQuestions and HybridQA~\citep{WikiTQ,HybridQA} assume that the relevant table is already given, and thus do not evaluate this retrieval problem. Recent Table RAG work~\citep{T-RAG,Pneuma,TableRAG} improves retrieval and reasoning, yet existing evaluations largely emphasize semantic relevance rather than whether a retrieved table is uniquely sufficient to answer the query. This makes it difficult to isolate answerability from coarse semantic matching.

To study this distinction, we introduce \textbf{\ours{}}, a controlled diagnostic benchmark for \textbf{T}able \textbf{C}ontent-level Answerability in \textbf{R}AG. \ours{} pairs answerable Target Tables with highly similar but non-answerable Sibling Distractors, varying answerability while keeping much semantic and structural context fixed. This controlled design tests whether a retriever can identify the uniquely answerable target rather than merely reaching its semantic neighborhood.

Using \ours{}, we identify a consistent \textbf{Semantic-Answerability Gap (SAG)} in retrieval: semantically relevant sources can be reached without reliably identifying the uniquely answerable one. We primarily study this phenomenon in \textbf{dense retrieval}. The best-performing dense retriever reaches the relevant table group at top-1 for 67.0\% of queries but retrieves the uniquely answerable target for only 18.2\%. This gap also has a clear downstream impact: QA F1 drops from 0.755 with the oracle table to 0.330 with retrieved top-5 tables. Diagnostics show that surface form is not the main cause: retrievers favor semantic volume over sufficiency, respond weakly to column-value binding disruptions, and form compact Target--Sibling neighborhoods without global collapse.

We further use \textbf{Answerability-Aware Reranking (AAR)} to test whether explicit query-conditioned answerability assessment can alleviate this failure. AAR improves top-1 retrieval from \textbf{18.2\%} to \textbf{57.4\%}. However, answerability-oriented fine-tuning is strongly distribution-dependent, generic retrieval training does not induce comparable answerability discrimination, and simple joint training does not recover the separate gains of the two objectives. Thus, learning answerability and preserving broad, transferable semantic retrieval remain distinct challenges.

\begin{figure}[t]
\centering
\includegraphics[width=0.99\linewidth]{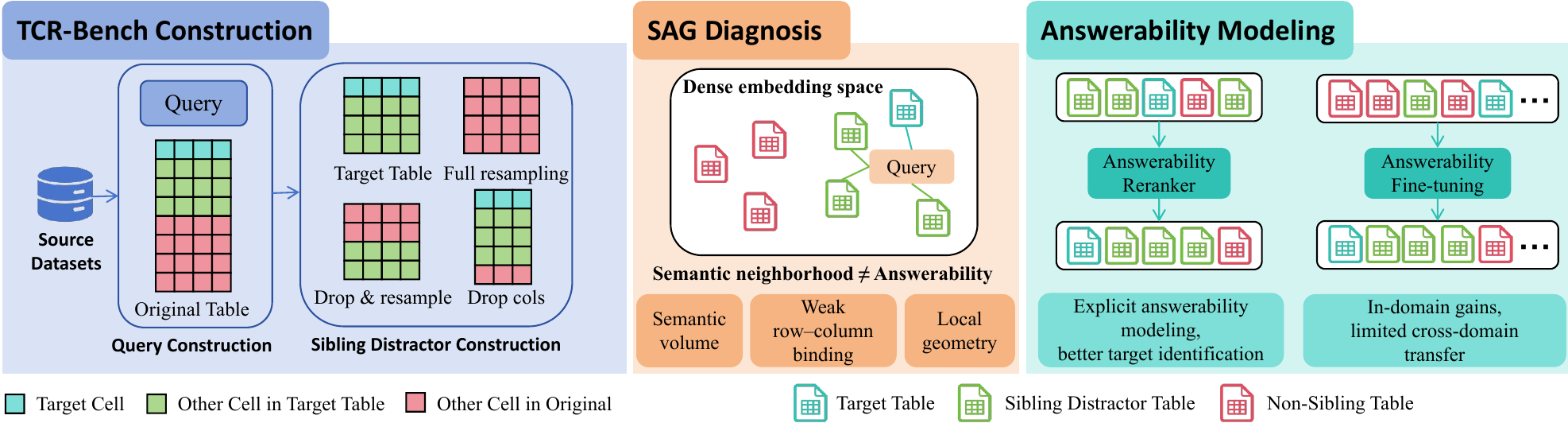}
\caption{Overview. Left: \ours{} pairs each answerable Target Table with semantically similar but non-answerable Sibling Distractors. Center: dense retrieval exhibits the \textbf{Semantic-Answerability Gap (SAG)}, with controlled diagnostics probing factors associated with this behavior. Right: explicit answerability modeling improves target identification, while answerability-oriented training shows limited cross-domain transfer.}
\label{fig:main}
\end{figure}

Figure~\ref{fig:main} summarizes the overall study. Our contributions are threefold.
\textbf{(1) We identify the Semantic-Answerability Gap:} we show that dense retrievers can reach the correct semantic neighborhood while failing to identify the uniquely answerable target, establishing answerability as a distinct retrieval dimension beyond semantic relevance.
\textbf{(2) We diagnose dense retrieval behavior in a controlled setting:} using \ours{}, we show that retrievers favor semantic volume over sufficiency, remain weakly sensitive to disrupted row-column bindings, and place Targets and Siblings in compact local neighborhoods without evidence of global collapse.
\textbf{(3) We investigate answerability-aware retrieval:} explicit answerability modeling substantially improves target identification, while answerability-oriented, generic retrieval, and joint training experiments show that transferable answerability discrimination and broad semantic retrieval do not necessarily align across settings.\footnote{Code and data are available at \url{https://anonymous.4open.science/r/TCR-Bench-05F6}.}

\section{Related Work}
\subsection{Beyond Semantic Relevance: Answerability in Retrieval}

Recent work has identified limitations of embedding-based retrieval beyond semantic similarity, including expressivity limits~\citep{embedding_theoretical_limitations}, false positives at scale~\citep{curse_of_dense}, diminishing returns from scaling~\citep{embedding_scaling_law}, and reliance on surface cues over factual evidence~\citep{collapse_dense_retrievers,fact_or_facsimile}.

More importantly, semantic similarity does not always imply answer support. Recent benchmarks show that similarity-based retrieval aligns more with topical relevance than decision usefulness~\citep{UsefulBench}, while answer-centric supervision improves retrieval when positives are defined by answer support~\citep{Beyond_Semantic_QA,ARK}. Other studies address misleading context~\citep{RAG-more-context2-Re-rag,Beyond_Semantic_Robust} and multi-hop retrieval~\citep{Beyond_Semantic_multi-hop1,Beyond_Semantic_multi-hop2}. However, these works do not isolate answerability as a distinct dimension from semantic relevance or systematically disentangle the two under controlled conditions. We formalize this mismatch as the \textbf{SAG} and study it through controlled sibling-table diagnostics.

\subsection{Table RAG Methods and Benchmarks}

Table retrieval has explored both table-specific dense retrievers \citep{NQ-TABLES-DTR} and generic retrievers such as DPR \citep{DPR}, with evidence that generic architectures can perform comparably to specialized ones~\citep{table-specific-embedding-not-better}. These findings suggest that structural bias alone may not fully capture fine-grained distinctions within tables. Meanwhile, existing evaluations have largely focused on semantic relevance, often leaving finer-grained content distinctions less examined. Simpler schema-aware and title-aware representations \citep{THoRR,TEM} have also shown strong performance, while hybrid and multi-table pipelines \citep{JAR,Mixture-of-RAG} similarly focus on relevance matching. Taken together, these settings often rely on relatively coarse signals of relevance, leaving finer-grained content-level distinctions less directly tested.

Existing benchmarks \citep{NQ-TABLES-DTR,WikiTables-RAG,TableCopilot,TARGET,t2-ragbench,mmRAG,MultiTableQA} increase difficulty through multi-hop reasoning, mixed modalities, or heterogeneous corpora, yet consistently frame retrieval as semantic matching. To our knowledge, existing benchmarks do not explicitly require distinguishing a uniquely answerable table from semantically similar candidates. \ours{} targets this gap directly through \textbf{sibling tables}, isolating fine-grained answerability selection from coarse semantic retrieval.

\section{A Controlled Setting for Studying Answerability}

\subsection{Answerability as a Distinct Retrieval Dimension}

We define \textbf{answerability} as whether a table contains sufficient information to produce a valid answer to a given query. This is distinct from semantic relevance: a table may be highly relevant to a query while still being unable to answer it. We therefore distinguish three table types: \textbf{Target Tables}, which are answerable; \textbf{Sibling Distractor Tables}, which are highly similar to the target but non-answerable; and \textbf{Non-Sibling Tables}, which are less related in schema or semantics.

\subsection{\ours{}: Controlled Sibling-Table Setting}

\ours{} is designed to diagnose the discrepancy between semantic relevance and answerability. For each query, we construct one Target Table and multiple Sibling Distractor Tables from the same source table. Sibling Distractor Tables preserve substantial semantic and structural similarity with the target while being modified to become non-answerable. This creates a controlled setting where semantic similarity is largely preserved while answerability varies.

We further vary queries, schemas, and table representations to reduce reliance on superficial matching. Tables are rendered in Markdown, HTML, CSV, and Mixed formats. Detailed construction and validation procedures are provided in Appendix~\ref{app:benchmark_detailed}.

\subsection{Benchmark Construction and Validation}

\ours{} is constructed from databases originating from Spider~\citep{Spider} and BIRD~\citep{BIRD}. We formulate three TableQA tasks with increasing complexity: \textbf{Exact Match (EM)}, \textbf{Selective Filtering (SF)}, and \textbf{Selective Aggregation (SA)}, involving 1--3 conjunctive conditions. Formal definitions are provided in Appendix~\ref{app:Task_Definitions}.

For each query, Sibling Distractor Tables are generated through controlled schema- and row-level modifications, followed by answerability verification and manual validation. We additionally define a diagnostic subset of EM queries with tables restricted to their sibling groups, allowing answerability to be examined under largely fixed semantic similarity. The final benchmark contains 209 queries (98 EM, 55 SF, and 56 SA) and 637 tables.

To test whether the observed phenomenon extends beyond controlled construction, we additionally introduce \textbf{Real-TCR}, a natural-negative collection derived from naturally occurring tables with real-world data partitions. Unlike \ours{}, its candidates are not generated through answerability-targeted transformations. Real-TCR contains 77 queries and 217 tables, while Real-TCR-small contains 38 queries and 117 tables. Detailed construction and results are provided in Appendix~\ref{app:real_tcr}.

\section{Semantic Relevance vs. Answerability}
\label{sec:main_results}

\begin{figure*}[t]
\centering
\includegraphics[width=0.9\linewidth]{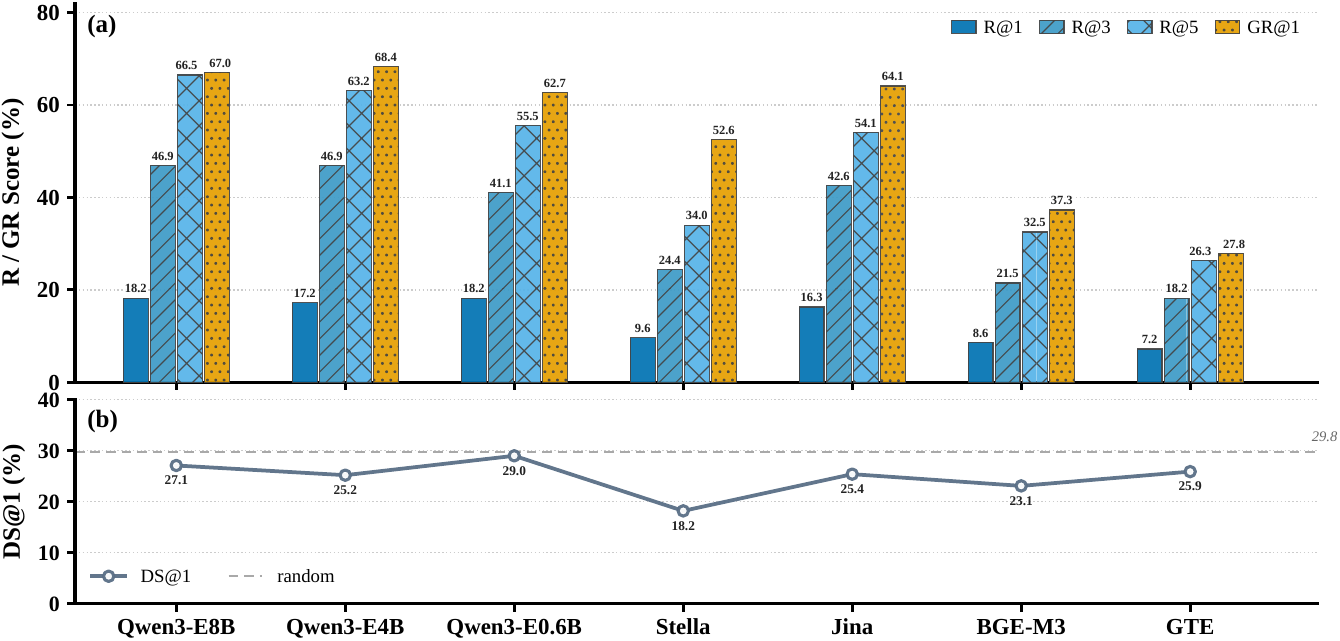}
\caption{Retrieval performance on \ours{} using the Mixed format. Abbreviated model names are used for compact presentation.}
\label{fig:main_result}
\end{figure*}

\subsection{Retrieval Evaluation}
\label{sec:main_results_R}

\subsubsection{Embedding Models}

We evaluate representative embedding models across architectural families and parameter scales, including Qwen3-Embedding-0.6B, Qwen3-Embedding-4B, Qwen3-Embedding-8B~\citep{qwen3embedding}, stella\_en\_1.5B\_v5~\citep{Stella}, jina-embeddings-v4~\citep{jina-embeddings-v4}, bge-m3~\citep{bge-m3}, and gte\_Qwen2-7B-instruct~\citep{gte-qwen}. Detailed model descriptions are provided in Appendix~\ref{app:embedding_models}.

\subsubsection{Evaluation Metrics}

We evaluate retrieval from two complementary perspectives: \emph{semantic relevance} and \emph{answerability}. Let $y_i$ be the ground-truth target and $R_i^k$ the top-$k$ retrieved items for query $i$.

\textbf{Top-$k$ Recall ($R@k$)}~\citep{TARGET, MultiTableQA, t2-ragbench} is defined as
$R@k=\frac{1}{N}\sum_{i=1}^{N}\mathbb{I}(y_i\in R_i^k)$.
Since each query in \ours{} has a unique answerable Target Table, $R@k$ measures whether the retriever successfully identifies the \textbf{answerable target}, rather than merely retrieving a semantically relevant table.

\textbf{Top-$k$ Group Recall ($GR@k$)} measures coarse semantic relevance. Let $\mathcal{G}_i$ denote the Sibling Table Group for query $i$, and $R_i^k$ the set of top-$k$ retrieved tables for query $i$. Inspired by the formulation of $CR@k$~\citep{metric-crk}, we define $GR@k=\frac{1}{N}\sum_{i=1}^{N}\frac{|R_i^k\cap\mathcal{G}_i|}{\min(|\mathcal{G}_i|,k)}$. We primarily report $GR@1$ to measure whether retrieval reaches the correct semantic neighborhood.

\textbf{Top-1 Discriminative Score ($DS@1$)} is defined as
$DS@1=R@1/GR@1$, with $DS@1=0$ when $GR@1=0$.
It measures how effectively the retriever identifies the answerable target within its semantically relevant group. Under random selection within a Sibling Table Group, the expected $DS@1$ is 0.221 on the original dataset and 0.298 when column-deletion variants are excluded.

\subsubsection{Main Results}

Figure~\ref{fig:main_result} reports retrieval performance under the \textbf{Mixed} format, with full results in Appendix~\ref{app:full_results}.

The results reveal a consistent \textbf{Semantic-Answerability Gap (SAG)} across the evaluated embedding models. The strongest dense retriever achieves $GR@1=0.670$ but only $R@1=0.182$, showing that identifying the correct semantic neighborhood is substantially easier than identifying the answerable target. Scaling Qwen3 from 0.6B to 8B improves $R@5$ by $19.8\%$, but yields only modest gains in $DS@1$.

Most models remain near or below the random-selection baseline of $DS@1=0.298$, with significance analysis in Appendix~\ref{sec:appendix_significance} confirming that they do not significantly outperform chance. Column-deletion cases provide a useful contrast: their $R@1$ drops to $0.072$ for Qwen3-E4B and $0.057$ for Stella, indicating sensitivity to explicit structural mismatches, while discrimination among schema-consistent tables remains difficult. These results suggest that dense retrieval captures topical and schema-level similarity effectively, but is substantially less sensitive to fine-grained content relations such as row-column bindings, value constraints, and condition consistency.

A consistent qualitative pattern emerges on Real-TCR. Jina achieves $GR@1=0.948$ but only $R@1=0.143$ and $DS@1=0.151$, while Qwen3-E4B reaches $GR@1=0.909$ with $DS@1=0.043$. Across both Real-TCR and Real-TCR-small, the evaluated retrievers consistently achieve high group recall but substantially lower target recall, indicating that the difficulty of distinguishing the uniquely answerable target persists beyond the controlled setting of the main benchmark. Complete results are reported in Appendix~\ref{app:real_tcr}.

More broadly, supplementary experiments with alternative first-stage retrievers, including DTR-style, ColBERT-style, sparse, and hybrid methods, show that the observed gap is not substantially reduced by these alternatives (Appendix~\ref{sec:appendix_alternative_retrieval} and~\ref{sec:appendix_augmentation}).

\subsection{Downstream QA Evaluation}
\label{sec:main_results_QA}

To examine how retrieval quality propagates to generation, we evaluate downstream QA using the LongTableBench~\citep{LongTableBench} pipeline with Qwen3-30B-A3B-Thinking-2507~\citep{qwen3technicalreport}. We report average F1 under five settings: Oracle (Target Table only), No-Table, and top-$k$ retrieval with $k\in\{1,3,5\}$, where retrieved tables are concatenated in retrieval order.

We define
$\text{Eff@}k=(\text{QA@}k/\text{QA@Oracle})/R@k$,
where $QA@k$ denotes QA F1 using top-$k$ retrieved tables. This measures how effectively the QA model exploits the Target Table once it is retrieved, with values near 1 indicating limited additional interference.

\begin{table}[htbp]
\centering
\caption{$QA@k$ and $Eff@k$ using top-$k$ retrieved tables from different embedding retrievers. The Oracle (Target Table only) and No-Table settings achieve F1 scores of 0.755 and 0.014, respectively.}
\label{tab:qa_results_full}
\begin{tabular}{l|ccc|ccc}
\toprule
\textbf{Method} & \textbf{QA@1} & \textbf{QA@3} & \textbf{QA@5} & \textbf{Eff@1} & \textbf{Eff@3} & \textbf{Eff@5} \\
\midrule
Qwen3-E0.6B & \textbf{0.144} & 0.244 & 0.276 & 1.046 & 0.786 & 0.659 \\
Qwen3-E4B   & 0.113 & \textbf{0.269} & 0.329 & 0.871 & 0.760 & 0.690 \\
Qwen3-E8B   & 0.139 & 0.266 & \textbf{0.330} & 1.014 & 0.751 & 0.657 \\
\bottomrule
\end{tabular}
\end{table}

The Oracle setting achieves $F1=0.755$, whereas No-Table drops to $0.014$, confirming that downstream answering depends heavily on retrieved tabular information. However, increasing $k$ provides limited benefit: even top-5 retrieval reaches only $F1=0.330$. Meanwhile, $Eff@1$ remains close to 1 across all three model sizes, indicating that QA is largely effective once the Target Table is ranked first. In contrast, $Eff@3$ and $Eff@5$ consistently decline as more tables are introduced, consistent with prior observations that additional semantically related context can cause interference in RAG systems~\citep{RAG-more-context1, RAG-more-context2-Re-rag, RAG-more-context3-lost-in-middle, RAG-more-context4}.

\subsection{The Answerability Bottleneck}
\label{sec:discussion}

The main experiments establish two observations. First, retrievers reliably reach the correct semantic group but rarely identify the answerable target. Second, downstream QA works well only when the target is ranked first, while deeper retrieval mostly adds semantically similar but non-answerable tables that interfere with generation.

\begin{tcolorbox}[colback=blue!5, colframe=black, fonttitle=\bfseries\small, fontupper=\small, boxrule=0.8pt, arc=2mm, left=2mm, right=2mm, top=1mm, bottom=1mm, title=The Answerability Bottleneck]
Dense retrievers identify semantically relevant sources well but often perform close to the random-selection baseline when distinguishing the answerable target within that group. This \textbf{SAG} persists across the models and scales we evaluate, including different retrieval settings.
\end{tcolorbox}

\section{Diagnosing Dense Retrieval Behavior}
\label{sec:diagnostic_analysis}

The main experiments show that dense retrieval can often reach the semantic neighborhood containing the Target Table, but has difficulty distinguishing it from closely related Sibling Distractors. We conduct controlled diagnostics to identify what information the evaluated representations rely on and where this discrimination failure arises. We first examine \emph{(1)} sensitivity to surface-level serialization and query wording, and then investigate \emph{(2)} sensitivity to semantic content and fine-grained table structure, including the embedding geometry underlying the observed behavior.

\subsection{Robustness to Surface Variations}
\label{sec:surface_variations}

We first test whether the observed behavior depends on how tables or queries are expressed. For tables, we evaluate six serialization formats on the Diagnostic Subset: CSV, HTML, Markdown, Mixed, and two sentence-based representations, \textbf{TSen} and \textbf{TSenS}. TSen converts each row into a declarative sentence while preserving attribute order, whereas TSenS randomly shuffles the attributes. For queries, we compare three semantically equivalent formulations: the original natural-language query, a sentential formulation, and a template-based formulation (Appendix~\ref{app:query-variants}).

Across these variations, retrieval remains broadly stable for both Qwen3-E4B and Stella. Qwen3-E4B achieves $R@1$ between 0.186 and 0.257 across the six table formats, with $GR@1$ between 0.757 and 0.857. Query reformulation similarly produces limited changes. Thus, the Target--Sibling discrimination failure is not primarily an artifact of serialization or query wording. Complete results and consistency analyses are provided in Appendix~\ref{app:surface_variation_results}.

\subsection{Sensitivity to Semantic and Structural Information}
\label{sec:retrieval_behavior}

We next examine which aspects of table content affect dense retrieval, focusing on semantic information volume and structural bindings relevant to answerability.

\paragraph{Semantic volume.}
We first ask whether dense retrieval responds to information useful for semantic matching or to information sufficient for identifying the answer. For each table row, we construct three sentence-based representations. \textbf{Sen} preserves the complete row and original attribute order. \textbf{SenS} preserves the complete row but randomly shuffles attributes, isolating sensitivity to ordering. \textbf{SenSS} retains only a subset of column-value pairs sufficient to uniquely identify the relevant row.

\begin{table}[htbp]
\centering
\small
\caption{Inter-table retrieval under different sentence-based representations. Sen preserves the complete row, SenS additionally shuffles the attribute order, and SenSS retains only a sufficient subset of attributes for answering the query.}
\label{tab:sentential_inter_app}
\begin{tabular}{l|l|ccc|c|c}
\toprule
\multirow{2}{*}{\textbf{Model}} & \multirow{2}{*}{\textbf{Type}} & \multicolumn{3}{c|}{\textbf{R@k}} & \multirow{2}{*}{\textbf{GR@1}} & \multirow{2}{*}{\textbf{DS@1}} \\
\cmidrule(lr){3-5}
 & & \textbf{@1} & \textbf{@3} & \textbf{@5} & & \\ 
\midrule
\multirow{3}{*}{Qwen3-E4B} 
& Sen & 0.232 & 0.529 & 0.800 & \textbf{0.987} & 0.235 \\
& SenS & \textbf{0.252} & \textbf{0.548} & \textbf{0.806} & \textbf{0.987} & \textbf{0.255} \\
& SenSS & 0.226 & 0.490 & 0.703 & 0.865 & 0.261 \\
\midrule
\multirow{3}{*}{Stella} 
& Sen & \textbf{0.213} & \textbf{0.561} & 0.768 & \textbf{0.994} & \textbf{0.214} \\
& SenS & 0.206 & 0.555 & \textbf{0.774} & 0.987 & 0.209 \\
& SenSS & 0.148 & 0.439 & 0.626 & 0.794 & 0.187 \\
\bottomrule
\end{tabular}
\end{table}

Table~\ref{tab:sentential_inter_app} shows that Sen and SenS perform similarly, whereas both generally outperform SenSS. For Qwen3-E4B, reducing a row to the sufficient subset lowers $R@5$ from 0.800--0.806 to 0.703. The small difference between Sen and SenS further indicates that attribute order is not the primary factor. Within these controlled representations, retrieval appears more responsive to the amount of semantic content than to whether the representation contains only information sufficient for identification.

\begin{figure}[htbp]
\centering
\includegraphics[width=0.95\linewidth]{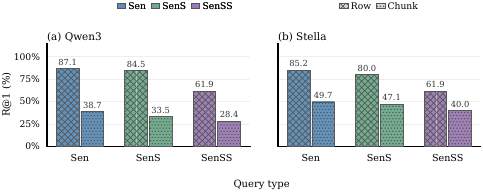}
\caption{Intra-table retrieval under different granularities and sentence-based representations. Row-level retrieval substantially outperforms 10-line chunks, while reducing a row to a sufficient subset in SenSS degrades retrieval.}
\label{fig:sentential_intra_full}
\end{figure}

Within the Target Table, row-level retrieval substantially outperforms retrieval over 10-line chunks (Figure~\ref{fig:sentential_intra_full}), indicating that dense embeddings capture local row-level content, while aggregation can dilute query-relevant information. Complete results are reported in Appendix~\ref{app:dense_retrieval_diagnostics}.

\paragraph{Row-column binding.}
Semantic content alone does not determine answerability: answers can depend on binding values to the correct columns and rows. On the Diagnostic Subset, we build a Minimal Contrastive Set with three configurations: \textbf{Ori} (original table), \textbf{RID} (each cell prefixed with its row index), and \textbf{RID+Shuf} (RID with values further shuffled within each column), which keeps explicit row indices and lexical content while breaking row--column bindings.

Table~\ref{tab:shuffle_results} shows that Ori and RID both outperform RID+Shuf (e.g., Qwen3-E4B Mixed: $0.443$/$0.371$ vs.\ $0.314$). Thus, even when row identity is made explicit, disrupting within-column value placements still degrades discrimination, indicating limited sensitivity to fine-grained row--column consistency. Full results are in Appendix~\ref{app:dense_retrieval_diagnostics}.

\begin{table}[htbp]
\centering
\small
\caption{Retrieval on the Minimal Contrastive Set. Ori is the original table; RID adds row indices; RID+Shuf further shuffles values within each column on top of RID.}
\label{tab:shuffle_results}
\begin{tabular}{l|l|c|c|c|c}
\toprule
\textbf{Model} & \textbf{Config} & \textbf{Format} & \textbf{R@1} & \textbf{GR@1} & \textbf{DS@1} \\
\midrule
\multirow{4}{*}{Qwen3-E4B} 
 & Ori      & Mixed    & 0.443 & 0.671 & 0.660 \\
 & Ori      & Markdown & 0.343 & 0.643 & 0.533 \\
 & RID      & Mixed    & 0.371 & 0.671 & 0.553 \\
 & RID+Shuf & Mixed    & 0.314 & 0.643 & 0.489 \\
\midrule
\multirow{3}{*}{Stella} 
 & Ori      & Mixed    & 0.229 & 0.471 & 0.485 \\
 & RID      & Mixed    & 0.243 & 0.471 & 0.515 \\
 & RID+Shuf & Mixed    & 0.200 & 0.429 & 0.467 \\
\bottomrule
\end{tabular}
\end{table}

\paragraph{Embedding geometry.}
The perturbation results show that the evaluated dense retrievers are more sensitive to semantic content than to some fine-grained structural changes. We next ask whether the discrimination failure reflects global representation collapse or local ambiguity among semantically related tables. Following~\citep{geometry_of_consolidation}, we report effective dimensionality $d_{\mathrm{eff}}$ (how many directions carry most of the variance) and mean pairwise cosine distance $\bar{d}$.

\begin{figure}[!ht]
\centering
\includegraphics[width=0.9\linewidth]{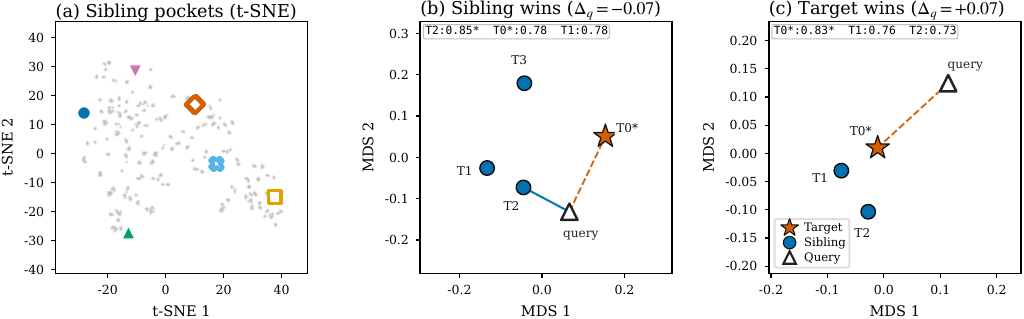}
\caption{Embedding geometry (Qwen3-E4B, Mixed).
(a)~Sibling groups form tight t-SNE pockets:
inn~(filled circle), baseball~(open square), WDI~(filled triangle),
music~(open diamond), codebase~(filled inverted triangle), games~(cross).
(b,c)~Within a pocket, cosine may prefer a Sibling ($\Delta_q{<}0$) or the Target ($\Delta_q{>}0$).}
\label{fig:embedding_geometry_main}
\end{figure}

We find evidence for the latter. Across all 637 tables, embeddings remain broadly dispersed ($d_{\mathrm{eff}}$ in the tens; $\bar{d}\in[0.40,0.55]$). In contrast, Target$+$Sibling groups form compact, low-dimensional neighborhoods (median $\bar{d}^{\mathrm{loc}}=0.08$--$0.11$; median $d_{\mathrm{eff}}^{\mathrm{loc}}\approx 2.4$--$2.9$; Table~\ref{tab:local_embedding_geometry_full}). Figure~\ref{fig:embedding_geometry_main} makes this concrete: sibling groups appear as tight pockets, while within-pocket cosine ranking can prefer a Sibling over the Target.

To quantify this final discrimination step, we use the Target--Sibling margin
$\Delta_q=\cos(q,t_q)-\max_{s\in S_q}\cos(q,s)$,
where $t_q$ is the Target and $S_q$ is the corresponding Sibling set. A positive margin means that the Target is closer to the query than every Sibling. Mean $\Delta_q$ is negative across all evaluated models and correlates more strongly with $R@1$ than with $GR@1$, indicating that remaining errors concentrate on query-to-table discrimination after the semantic neighborhood has been reached. Full analyses are in Appendix~\ref{app:embedding_geometry}.

\paragraph{Diagnostic Summary.}
Within our evaluation, serialization, query wording, and global embedding collapse do not appear to explain the observed gap. Instead, the tested dense retrievers tend to favor semantic volume over information sufficiency, show weak sensitivity to row-column bindings, and place Targets and Siblings in compact local neighborhoods that remain difficult to separate.

\begin{tcolorbox}[colback=blue!5, colframe=black, fonttitle=\bfseries\small, fontupper=\small, boxrule=0.8pt, arc=2mm, left=2mm, right=2mm, top=1mm, bottom=1mm, title=Similarity Is Not Sufficiency]
The gap is not a surface artifact: dense retrievers favor semantic volume over sufficiency, respond weakly to row-column bindings, and leave Targets and Siblings hard to separate in compact local neighborhoods.
\end{tcolorbox}

\section{Modeling Answerability for Retrieval}
\label{sec:answerability_modeling}

To further examine whether explicitly modeling answerability can alleviate the SAG, we conduct two complementary experiments: answerability-aware reranking and answerability-oriented fine-tuning. The former asks whether query-conditioned interaction can distinguish the target from semantically similar tables, while the latter asks whether this behavior can be learned by an embedding model and transferred beyond the training distribution.

\paragraph{Query-Conditioned Answerability Assessment.}
We use \textbf{Answerability-Aware Reranking} (\textbf{AAR}) as a diagnostic framework to explicitly assess answerability after dense retrieval. Unlike single-vector retrieval, query-conditioned interaction directly evaluates whether a candidate table is sufficient to satisfy the query. We consider three variants: \textbf{AAR-CE}, using Qwen3-Reranker-8B (Qwen3-R8B) paired with the Qwen3-E8B retriever; \textbf{AAR-LLM-PW}, using Qwen3-30B as a pointwise LLM-based answerability classifier~\citep{Pneuma}; and \textbf{AAR-LLM-LW}, using the same Qwen3-30B model to jointly rank candidates in a listwise setting.

\begin{table}[htbp]
\centering
\small
\caption{Answerability-aware reranking on the Qwen3-E8B backbone ($k{=}10$).}
\label{tab:qwen3-emb-8b-results}
\begin{tabular}{l|cc|c|c}
\toprule
\textbf{Method} & \textbf{R@1} & \textbf{R@5} & \textbf{GR@1} & \textbf{DS@1} \\
\midrule
AAR-CE & \textbf{0.574} & 0.656 & 0.766 & \textbf{0.750} \\
AAR-LLM-PW & 0.536 & \textbf{0.737} & 0.761 & 0.704 \\
AAR-LLM-LW & 0.450 & 0.708 & \textbf{0.770} & 0.584 \\
\bottomrule
\end{tabular}
\end{table}

All three variants substantially improve target identification over dense retrieval, with R@1 increasing from 0.182 to 0.450--0.574. The gain persists under both pointwise and listwise LLM reranking, indicating that it is not specific to a particular reranking formulation. The matched Qwen3-E8B and Qwen3-R8B comparison further suggests that the improvement is associated with the change from dense embedding to query-conditioned reranking rather than model scale: both use the same pretrained initialization and 8B parameter scale, but differ in objective. Full results and significance analysis are in Appendix~\ref{app:full_results_rerank} and Appendix~\ref{sec:appendix_significance}.

\paragraph{Learning Answerability in Embeddings.}
We further fine-tune Qwen3-E0.6B with query--target positives and sibling-table hard negatives to test whether answerability can be learned directly by an embedding model. The best TCR-Bench model improves R@1 from 0.182 to 0.493 and DS@1 from 0.290 to 0.691, showing that answerability is learnable under matched supervision. However, the learned signal is strongly distribution-dependent: the TCR-Bench-trained model achieves only 0.065 R@1 on Real-TCR and substantially degrades retrieval on the Table RAG benchmark TARGET~\citep{TARGET}. The reverse transfer from Real-TCR to TCR-Bench is similarly weak, while generic training on TabFact and FeTaQA, two subsets of TARGET, improves semantic retrieval without inducing comparable answerability-sensitive discrimination. Moreover, a simple mixture of TCR-Bench and TARGET training data does not simultaneously recover the gains of the two objectives. Thus, learning answerability and preserving broad, transferable semantic retrieval remain distinct challenges. Detailed configurations and results are given in Appendix~\ref{app:answerability_training}.

\begin{tcolorbox}[colback=blue!5, colframe=black, fonttitle=\bfseries\small, fontupper=\small, boxrule=0.8pt, arc=2mm, left=2mm, right=2mm, top=1mm, bottom=1mm, title={Answerability Is Learnable, but Transfer Remains Difficult}]
Explicit answerability modeling improves target identification in our evaluation, but learning transferable answerability discrimination while preserving broad semantic retrieval remains nontrivial.
\end{tcolorbox}

\section{Conclusion}
Our experiments reveal a consistent \textbf{Semantic-Answerability Gap (SAG)} in retrieval: semantically relevant sources can be reached without reliably identifying the uniquely answerable target. We primarily study this phenomenon in \textbf{dense retrieval}, using \textbf{\ours{}} to isolate the gap through sibling tables that are highly similar in semantics and structure but differ in answerability. Our diagnostics show that retrievers favor semantic volume over sufficiency, respond weakly to row-column bindings, and leave Targets and Siblings hard to separate locally. Explicit answerability modeling substantially improves target identification, while fine-tuning shows that answerability can be learned but remains strongly distribution-dependent. Generic retrieval training does not automatically induce comparable answerability-sensitive discrimination, and simple joint training does not simultaneously preserve gains in answerability and broad semantic retrieval. These findings motivate retrieval objectives and models that treat answerability as a distinct retrieval dimension while jointly preserving broad semantic relevance and transferable answerability discrimination.

\subsection*{AI use statement}

In this work, we used generative AI tools for minor code generation assistance, manuscript editing and polishing, and paraphrasing queries and table schemas during dataset construction. We have not used generative AI tools for developing theoretical models or conceptual frameworks, formulating mathematical claims, assisting in proofs, proposing or refining hypotheses, designing or providing feedback on research methodology or experiments, cleaning or reformatting datasets, performing qualitative or thematic data analysis, or interpreting experimental results, and the remaining required disclosure tasks are not applicable to this work. Additionally, we used generative AI tools for retrieving and discovering information, such as finding related work. We have reviewed all AI-assisted work. LLM-paraphrased queries and table schemas were manually verified by the authors for semantic correctness and appropriateness, LLM-generated code was tested for correctness before use, and all AI-assisted manuscript edits were reviewed and revised by the authors. We take responsibility for the final content of this work, including text, claims, and artifacts produced with the aid of generative AI.

\subsection*{Ethics statement}

\ours{} is constructed from publicly available databases, including Spider and BIRD, together with synthetically generated table variants and query paraphrases. The benchmark contains no personally identifiable information. Dataset construction involves automatic generation followed by manual author verification to check the quality and validity of queries and table variants. The benchmark and accompanying resources are intended for research use under the stated license.

As a supplementary natural-negative validation, we additionally evaluate \textbf{Real-TCR}, a small collection derived from de-identified operational tables provided under a data-use agreement with an industry partner and annotated by personnel of that organization. Due to contractual and data-use constraints, Real-TCR is not publicly released, including in de-identified form. Real-TCR is used only for external validation and cross-distribution diagnostics and is not required to reproduce the main findings or controlled analyses of \ours{}. We do not release any underlying operational data or identifying information from this collection. Practitioners applying Table RAG methods to sensitive or proprietary data should use appropriate access controls, data governance, and privacy safeguards beyond those provided by this benchmark.

\subsection*{Reproducibility statement}

The code and data for \ours{} are publicly available at \url{https://github.com/minger-hsxz/TCR-Bench-Open}. The repository provides the core resources needed to reproduce the main experiments and controlled diagnostics, including the table data, complete query files, the Diagnostic subset, the RAG pipeline implementation, and configuration files specifying the experimental settings and hyperparameters. The implementation covers dense retrieval, AAR reranking, and downstream TableQA used in the main experiments. The paper and appendix provide the benchmark construction procedure, evaluation metrics, experimental settings, and additional diagnostic analyses needed to reproduce and interpret the reported results. Dense retrieval components produce deterministic outputs given fixed model weights and inputs. LLM-based components are run with temperature set to 0, although minor nondeterminism may remain due to hardware, software, or library implementations.

As described in the Ethics Statement, Real-TCR cannot be publicly released because of data-use restrictions on the underlying industry data. Consequently, results that directly depend on Real-TCR cannot be independently reproduced from the provided resources. These results serve as supplementary external validation, while the main \ours{} results and controlled analyses are reproducible using the publicly released resources.



\bibliography{iclr2027_conference}
\bibliographystyle{iclr2027_conference}

\appendix
\section{Benchmark Construction Details}
\label{app:benchmark_detailed}

This section provides the detailed construction procedures and implementation constraints used to instantiate \ours{}. The goal is to isolate content-level answerability from coarse semantic relevance: for each query, an answerable Target Table is paired with highly similar but non-answerable Sibling Distractor Tables, so that answerability can vary while much of the semantic and structural context remains fixed. Here we describe how these tables are generated, how surface-level shortcuts are controlled, and how unique answerability and annotation quality are verified.

\subsection{Source Tables and Query Generation}
\label{app:source_query}

We construct \ours{} from databases originating from Spider~\citep{Spider} and BIRD~\citep{BIRD}. These benchmarks provide relational tables from diverse real-world domains and naturally occurring schemas, allowing our controlled construction to retain realistic table structures while enabling precise manipulation of answer-bearing evidence.

For each source table, we generate TableQA queries based on its schema and cell values. We consider three task types: Exact Match (EM), Selective Filtering (SF), and Selective Aggregation (SA), corresponding to progressively more complex query semantics. Each query contains 1--3 conjunctive conditions over target columns. The formal definitions and examples of these tasks are provided in Appendix~\ref{app:Task_Definitions}.

For each query, the original source table serves as the basis for constructing a Target Table and its associated Sibling Distractor Tables. The Target Table remains uniquely answerable, whereas Sibling Distractors are produced by controlled schema- and row-level modifications that preserve substantial semantic and structural similarity while removing the evidence needed to answer the query.

\subsection{Task Definitions}
\label{app:Task_Definitions}

We formalize three query-answer tasks of increasing semantic complexity: Exact Match (EM), Selective Filtering (SF), and Selective Aggregation (SA). Each task is defined over a relational table $\mathcal{T}$ with schema $\mathcal{S}=(A_1,\dots,A_k)$, where each attribute $A_i$ takes values from a discrete or continuous domain.

A query $Q$ comprises a set of selection predicates $\mathcal{P}$ and, depending on the task type, may additionally specify a target attribute $A_t$ and an aggregation function $f$. Let $R\subseteq\mathcal{T}$ denote the set of rows satisfying all predicates in $\mathcal{P}$:

$$
R=\{\,r\in\mathcal{T}\mid \forall p\in\mathcal{P}:p(r)=\text{true}\,\}.
$$

\paragraph{Exact Match (EM).}

In the EM setting, every predicate in $\mathcal{P}$ is a conjunction of equality conditions of the form $A_i=v_i$, where $v_i$ is a literal value from the domain of $A_i$. The query requests the value(s) of a designated target column $A_t$ from the rows in $R$. Under the key-based uniqueness assumption, $R$ is either empty or a singleton. In practice, the answer is the unique value of $A_t$ in the matching row.

\textit{Example.} For the table shown in Table~\ref{tab:example_table}, the query ``What is C when A = 5 and B = 3?'' yields the unique answer $4$.

\paragraph{Selective Filtering (SF).}

The SF task relaxes the equality restriction. At least one predicate employs a comparison operator from the set
${<,>,\leq,\geq,\neq}$. Predicates may still contain equality conditions, but the filter is no longer restricted to exact key lookups. The answer is the set of values of the target column $A_t$ over the filtered rows $R$.

\textit{Example.} The query ``What is C when A = 5 and B < 5?'' filters the rows using an equality condition on $A$ and an inequality condition on $B$, returning $4$.

\paragraph{Selective Aggregation (SA).}

The SA task extends SF with a downstream aggregation operation. The query first applies the selection predicates $\mathcal{P}$ to obtain the filtered set $R$, and then computes an aggregate statistic over the values of a target column $A_t$. The aggregation function $f$ is drawn from
${\text{SUM},\text{AVG},\text{MIN},\text{MAX}}$.

The answer is the scalar result

$$
f(\{\,r[A_t]\mid r\in R\,\}).
$$

\textit{Example.} The query ``What is the average of C when A $\geq$ 5 and B $\geq$ 3?'' filters the rows satisfying both inequalities and returns the mean of the corresponding C values, which is $5$.

\begin{table}[h!]
\centering
\caption{Illustrative relational table for the task definitions.}
\label{tab:example_table}
\begin{tabular}{|c|c|c|}
\hline
A & B & C \\
\hline
5 & 3 & 4 \\
6 & 8 & 6 \\
\hline
\end{tabular}
\end{table}

These three tasks form a progression from exact symbolic lookup to conditional filtering and aggregation over a set of retrieved values. Together, they probe complementary reasoning requirements for determining whether a table contains sufficient evidence to answer a query.

\subsection{Sibling Distractor Generation}
\label{app:sibling_generation}

The primary objective of Sibling Distractor construction is to preserve semantic and structural similarity while removing answer-bearing evidence. We therefore generate distractors from the same source table rather than independently sampling unrelated tables.

We employ three complementary generation strategies:

\begin{itemize}
\item \textbf{Removing Condition or Target Columns.} We remove either a query-condition column or the target column from the source table. Only one column is removed at a time, so that the resulting table retains most of the original schema while becoming insufficient for answering the query.

\item \textbf{Row Deletion.} We randomly delete 30\%--70\% of the rows from the source table. Additional rows are subsequently sampled from the remaining corpus when necessary to satisfy the predefined table-length constraint. Rows that would provide the required answer evidence are explicitly excluded.

\item \textbf{Row Resampling.} We construct a new table by resampling rows from the available table corpus while preserving the relevant schema characteristics. The resulting table remains structurally plausible but does not contain the evidence required by the query.

\end{itemize}

These strategies create distractors with different forms of missing evidence. Column-level modifications primarily remove schema or attribute grounding, whereas row-level modifications remove the specific records required to satisfy the query. Combining these mechanisms prevents the benchmark from reducing answerability to a single perturbation pattern.

\subsection{Sampling Constraints and Answer Filtering}
\label{app:sampling_constraints}

To ensure comparable retrieval conditions across candidate tables, all generated sub-tables are constrained to approximately 6,500--7,800 tokens, measured using \texttt{tiktoken}\footnote{\url{https://github.com/openai/tiktoken}}. This relatively large context size is intended to embed the relevant evidence within a substantial amount of table content, so that retrieval cannot be trivially solved by inspecting a small table. As we further show in Appendix~\ref{sec:appendix_length_scaling}, however, this challenge persists even for smaller tables.

For row deletion and row resampling, we explicitly filter out rows that would allow the query to be answered correctly. More importantly, answerability is verified globally across the complete benchmark corpus rather than only within the corresponding sibling group. This prevents an independently sampled table from accidentally becoming an alternative answerable table for the same query.

After this filtering procedure, each query is associated with exactly one valid Target Table. Sibling Distractor Tables remain semantically plausible but fail at least one requirement necessary for producing a valid answer.

\subsection{Query and Schema Augmentation}
\label{app:augmentation}

We introduce controlled semantic and surface variations to reduce the possibility that retrieval can be solved through direct lexical matching. These transformations are applied while preserving the underlying query semantics and answerability relation. Tables are additionally rendered in Markdown, HTML, CSV, and Mixed formats. The Mixed format refers to a combination of the first three, where each table is randomly assigned one of them; this assignment is pre-determined to ensure experimental consistency.

\subsubsection{Query and Schema Paraphrasing}

Queries and schemas are paraphrased to introduce linguistic and naming variation. The generated paraphrases are required to preserve the original query intent, constraints, and target attributes. Manual validation is subsequently applied to cases where the generated text may introduce semantic shifts.

Schema paraphrasing is particularly important because Sibling Distractor Tables originate from the same source table and therefore naturally share substantial lexical overlap. Varying column names reduces the extent to which retrieval can rely on direct schema-token matching.

\subsubsection{Date Format Augmentation}

Date values are transformed into multiple surface representations while preserving their underlying temporal values. Specifically, standard $YYYY-MM-DD$ representations may be converted into:

\begin{itemize}
\item \textbf{English Abbreviations:} e.g., \textit{15/Jan/2023} or \textit{Jan. 15, 2023}.
\item \textbf{Full English Spellings:} e.g., \textit{January 15th, 2023}.
\item \textbf{Roman Numerals:} e.g., \textit{2023-I-15}.
\item \textbf{Numeric Words:} e.g., \textit{two thousand twenty-three. one. fifteen}.
\end{itemize}

These transformations preserve the semantic identity of dates while substantially changing their token-level representations.

\subsubsection{Numeric and Scale Perturbations}

We similarly transform numerical values to reduce reliance on exact lexical matching. The augmentation strategies include:

\begin{itemize}
\item \textbf{Scaling Large Numbers:} numbers in the thousands are represented using compact forms with \textit{k} or \textit{K}, e.g., $15000 \rightarrow 15k$.

\item \textbf{Multiplicative and Additive Shifts:} numerical values are perturbed through controlled multiplication, division, or small additive shifts. The type of perturbation applied to each column is explicitly indicated in the column name and also communicated to the model in the relevant prompt. This type of perturbation is applied relatively infrequently.

\item \textbf{Numeral-System Conversion:} numbers are represented using Roman numerals or English words, e.g., $102 \rightarrow$ \textit{C} or \textit{One hundred and two}.

\end{itemize}

All transformations are applied subject to the requirement that they do not alter the intended query semantics or introduce unintended answerable distractors.

\subsection{Semantic Augmentation Model}
\label{app:gpt-oss}

We use GPT-OSS-120B~\citep{gpt-oss-120b}, an open-weight model released by OpenAI under the Apache 2.0 license, for query and schema paraphrasing. At the time of benchmark construction, it provided a strong locally deployable open-weight option, allowing the augmentation process to be reproduced without reliance on proprietary API endpoints.

The model is used only for semantic augmentation and does not participate in retrieval or downstream question answering. To reduce the possibility that benchmark behavior is tied to the augmentation model, subsequent experimental components, including dense retrievers, rerankers, and QA models, are selected from non-GPT model families.

All generated paraphrases are subsequently validated to ensure that they preserve the original query intent and schema semantics.

\subsection{Manual Validation and False-Negative Reduction}
\label{app:manual_validation}

We employ a multi-stage validation procedure to improve annotation quality and reduce false-negative cases.

First, answerability is verified programmatically across all tables. Within each sibling group, only the Target Table is allowed to satisfy all query constraints. We additionally evaluate the query logic against the full candidate corpus to identify tables outside the sibling group that may inadvertently contain sufficient evidence to answer the query.

Second, we conduct targeted manual inspection of potentially ambiguous tables. For each query, we retrieve the top-$k$ candidate tables outside its corresponding sibling group using multiple similarity signals. Candidate sets from different retrieval signals are merged, and the resulting tables are manually examined for semantic equivalence with the Target Table. This procedure focuses on cases where different schemas or column names may express the same underlying concept and therefore create unintended alternative answer tables. Ambiguous cases are removed from the benchmark.

Third, all generated query and schema transformations are manually checked. Annotators verify that rewritten queries preserve the original intent and constraints, and that schema transformations remain semantically consistent with the source tables.

Approximately 5\% of GPT-OSS-generated paraphrases required manual correction due to semantic shifts.

Although complete elimination of false negatives is difficult in an open-ended semantic space, the combination of global answerability checking, targeted retrieval-based inspection, and manual validation substantially reduces this risk.

\subsection{Benchmark Statistics}
\label{app:benchmark_statistics}

The final benchmark contains 209 queries, including 98 EM, 55 SF, and 56 SA queries, together with 637 tables. Tables may be shared across different query instances. We additionally define a diagnostic subset restricted to sibling groups, allowing answerability to be examined under largely fixed semantic similarity.

The benchmark is intentionally moderate in corpus size because its primary purpose is controlled diagnosis rather than large-scale corpus retrieval. The difficulty is introduced through the construction of semantically similar Sibling Distractor Tables, unique answerability, large table contexts, and surface-level variations.


\section{Detailed Description of Embedding Models}
\label{app:embedding_models}

This appendix summarizes the embedding model families evaluated in the main experiments. Since many table-specific embedding models, like DTR~\citep{NQ-TABLES-DTR}, often use very limited context, and studies such as Wang et al.~\citep{table-specific-embedding-not-better} have shown that table-specific embeddings do not necessarily outperform general text embeddings in the Table RAG domain, we focus only on commonly used, open-source, and relatively strong embedding models, all of which are capable of handling 8K context windows.

\begin{itemize}

\item \textbf{Qwen3 Embedding Family~\citep{qwen3embedding}.}  
The \textbf{Qwen3} embedding series includes \textbf{Qwen3-E0.6B}, \textbf{Qwen3-E4B}, and \textbf{Qwen3-E8B}. These instruction-aware text embedding models support long contexts (up to 32k tokens), multilingual input (100+ languages), and Matryoshka Representation Learning (MRL) for flexible embedding dimensions. Larger variants generally provide stronger semantic representation quality.

\item \textbf{Stella~\citep{Stella}.}  
\textbf{stella\_en\_1.5B\_v5} is a 1.5B-parameter English sentence embedding model optimized for semantic similarity and dense retrieval. It produces fixed-length embeddings and serves as a competitive mid-scale dense baseline.

\item \textbf{Jina~\citep{jina-embeddings-v4}.} 
\textbf{jina-embeddings-v4} is a $\sim$3.8B-parameter multilingual and multimodal embedding model. It supports unified text-image representations, as well as both single-vector dense and multi-vector late-interaction retrieval.

\item \textbf{BGE-M3~\citep{bge-m3}.} 
\textbf{bge-m3} is a multilingual embedding model designed for dense, sparse (lexical), and multi-vector retrieval within a unified architecture. It supports long inputs (up to 8192 tokens) and enables hybrid retrieval pipelines.

\item \textbf{GTE~\citep{gte-qwen}.} 
\textbf{gte\_Qwen2-7B-instruct} is a 7B-parameter instruction-tuned embedding model from the GTE series. It is optimized for semantic matching and large-scale dense retrieval.

\end{itemize}

\textbf{Model Diversity.} 
The selected models differ in parameter scale (0.6B–8B+), modality support (text-only vs. multimodal), and retrieval design (dense-only vs. hybrid/multi-vector); however, we utilize only text-only dense retrieval in this work.

\newtcolorbox{promptbox}[1]{
    colback=gray!5!white,
    colframe=gray!75!black,
    fonttitle=\bfseries,
    title=#1,
    boxrule=0.5pt,
    arc=2pt,
    left=5pt,
    right=5pt,
    top=5pt,
    bottom=5pt,
    standard jigsaw,
    opacityback=0.5
}

\section{Prompts Used in Our Pipeline}

\subsection{Retrieval Prompt}
Since we utilize the Qwen3-Embedding series, which is an instruction-aware retrieval model, we prepend a specific instruction to the user query to guide the embedding space towards TableQA relevance. The exact instruction used is detailed in Table~\ref{tab:retrieval-prompt}. Note that traditional lexical methods, such as BM25~\citep{BM25}, do not utilize this instruction.

\begin{table*}[t]
\caption{Instruction prepended to user queries for the retrieval stage.}
\label{tab:retrieval-prompt}
\begin{promptbox}{Retrieval Instructions}
\small
\texttt{Given a TableQA query, retrieve a relevant table that can answer the query. Note that the retrieved table should contain sufficient information to provide an answer, rather than resulting in an empty answer. [Context-Specific Task Description]}
\end{promptbox}
\end{table*}

\subsection{Downstream Question Answering Prompt}
For the RAG-based generation stage, we adopt a structured prompt following the style of LongTableBench. This prompt enforces strict formatting constraints, including numerical normalization (Roman to Arabic), handling of derived columns (e.g., reversing mathematical operations), and date standardization.

\begin{table*}[t]
\caption{The structured prompt used for the downstream TableQA generation stage.}
\label{tab:qa-prompt}
\begin{promptbox}{QA System Prompt}
\small
\textbf{System Prompt:} \\
\#\#\# Requirements: \\
Please read the following table and then answer the questions based on the table. Organize your answers into a list of strings, with each element being an answer item (The number of answer items is less than 11). If the question includes a special requirement, such as outputting a dictionary, please fulfill that specific request. Otherwise, always output a list of strings, even if there is only one answer item. Please place your answers between the \verb|```| and \verb|```|. \\

\smallskip
\#\#\# Notes: \\
The table may include non-standard formats for numbers or dates. For numbers, formats may include Roman numerals, English words, or scientific notation. Whenever possible, please convert these into Arabic numerals in your response. \\
Additionally, some columns are derived through mathematical operations. For example, \texttt{total\_add\_10} indicates that the values in this column are obtained by adding 10 to the original values. You should return the original values (i.e., subtract 10 from the current values). Furthermore, for any values ending with ``k'' or ``K'', convert them into standard Arabic numerals. Please convert all dates to the format \%Y-\%m-\%d <other time elements> (in Arabic numerals) in your response. \\

\smallskip
Additionally, even if there are duplicate answer items, you need to output all of them. \\

\smallskip
\# Example Output Format: \\
\verb|```|[`answer\_item\_1', `answer\_item\_2', ..., `answer\_item\_n']\verb|```| \\

\smallskip
Please do not generate any text after outputting the final answer.

\medskip
\hrule
\medskip

\textbf{User Prompt:} \\
table information: \\
\texttt{\{table\_infos\}} \\

Question: \texttt{\{query\}} \\
Answer:
\end{promptbox}
\end{table*}

As shown in Table~\ref{tab:qa-prompt}, the system prompt ensures that the model outputs are parsable and normalized across diverse table schemas.

\section{Comprehensive Retrieval Results}

This section collects the full dense-retrieval tables and related first-stage comparisons that support Section~\ref{sec:main_results}, including alternative paradigms, query-type breakdowns, and table-length scaling.

\subsection{Dense Embedding Retrieval Results}
\label{app:full_results}
Table~\ref{tab:full_results} presents the complete benchmark results for all evaluated models across all supported formats.

\begin{table*}[tp]
\centering
\caption{Comprehensive evaluation results across different input formats.}
\label{tab:full_results}
\resizebox{\textwidth}{!}{
\begin{tabular}{l|l|ccccc|ccccc|c}
\toprule
\multirow{2}{*}{\textbf{Model}} & \multirow{2}{*}{\textbf{Format}} & \multicolumn{5}{c|}{\textbf{R@k}} & \multicolumn{5}{c|}{\textbf{GR@k}} & \multirow{2}{*}{\textbf{DS@1}} \\
\cmidrule(lr){3-7} \cmidrule(lr){8-12}
& & \textbf{@1} & \textbf{@2} & \textbf{@3} & \textbf{@4} & \textbf{@5} & \textbf{@1} & \textbf{@2} & \textbf{@3} & \textbf{@4} & \textbf{@5} & \\
\midrule
\multirow{6}{*}{Qwen3-Embedding-0.6B} & CSV & 0.139 & 0.273 & 0.378 & 0.469 & 0.526 & 0.608 & 0.605 & 0.579 & 0.561 & 0.539 & 0.228 \\
 & Mixed & 0.182 & 0.297 & 0.411 & 0.502 & 0.555 & 0.627 & 0.617 & 0.593 & 0.574 & 0.556 & 0.290 \\
 & HTML & 0.182 & 0.282 & 0.402 & 0.478 & 0.574 & 0.651 & 0.636 & 0.616 & 0.590 & 0.583 & 0.279 \\
 & Markdown & 0.153 & 0.306 & 0.383 & 0.455 & 0.555 & 0.603 & 0.596 & 0.573 & 0.548 & 0.532 & 0.254 \\
 & TSen & 0.144 & 0.297 & 0.397 & 0.478 & 0.545 & 0.641 & 0.624 & 0.596 & 0.573 & 0.560 & 0.224 \\
 & TSenS & 0.124 & 0.287 & 0.373 & 0.483 & 0.541 & 0.641 & 0.622 & 0.585 & 0.567 & 0.559 & 0.194 \\
\midrule
\multirow{6}{*}{Qwen3-Embedding-4B} & CSV & 0.158 & 0.311 & 0.464 & 0.579 & 0.646 & 0.656 & 0.682 & 0.681 & 0.663 & 0.653 & 0.241 \\
 & Mixed & 0.172 & 0.321 & 0.469 & 0.560 & 0.632 & 0.684 & 0.708 & 0.686 & 0.670 & 0.662 & 0.252 \\
 & HTML & 0.134 & 0.287 & 0.474 & 0.584 & 0.679 & \textbf{0.713} & \textbf{0.725} & \textbf{0.708} & \textbf{0.696} & \textbf{0.687} & 0.188 \\
 & Markdown & 0.187 & 0.340 & \textbf{0.512} & 0.569 & 0.656 & 0.703 & 0.706 & 0.691 & 0.682 & 0.666 & 0.265 \\
 & TSen & 0.139 & 0.306 & 0.440 & 0.545 & 0.632 & 0.684 & 0.682 & 0.667 & 0.656 & 0.653 & 0.203 \\
 & TSenS & 0.158 & 0.321 & 0.488 & 0.579 & 0.646 & 0.684 & 0.667 & 0.663 & 0.652 & 0.644 & 0.231 \\
\midrule
\multirow{6}{*}{Qwen3-Embedding-8B} & CSV & 0.124 & 0.301 & 0.445 & 0.545 & 0.627 & 0.646 & 0.636 & 0.632 & 0.611 & 0.610 & 0.193 \\
 & Mixed & 0.182 & 0.344 & 0.469 & \textbf{0.608} & 0.665 & 0.670 & 0.658 & 0.641 & 0.628 & 0.622 & 0.271 \\
 & HTML & 0.153 & 0.335 & 0.488 & 0.598 & \textbf{0.684} & 0.699 & 0.689 & 0.683 & 0.679 & 0.674 & 0.219 \\
 & Markdown & 0.177 & 0.349 & 0.493 & 0.565 & 0.632 & 0.660 & 0.648 & 0.646 & 0.634 & 0.629 & 0.268 \\
 & TSen & 0.148 & 0.297 & 0.431 & 0.531 & 0.593 & 0.636 & 0.639 & 0.611 & 0.603 & 0.596 & 0.233 \\
 & TSenS & 0.177 & 0.306 & 0.407 & 0.507 & 0.560 & 0.632 & 0.632 & 0.619 & 0.604 & 0.597 & 0.280 \\
\midrule
\multirow{6}{*}{bge-m3} & CSV & 0.067 & 0.177 & 0.273 & 0.354 & 0.388 & 0.459 & 0.419 & 0.402 & 0.384 & 0.376 & 0.146 \\
 & Mixed & 0.086 & 0.177 & 0.215 & 0.282 & 0.325 & 0.373 & 0.352 & 0.338 & 0.334 & 0.316 & 0.231 \\
 & HTML & 0.043 & 0.139 & 0.225 & 0.263 & 0.292 & 0.349 & 0.347 & 0.346 & 0.341 & 0.330 & 0.123 \\
 & Markdown & 0.081 & 0.148 & 0.244 & 0.297 & 0.344 & 0.364 & 0.356 & 0.365 & 0.349 & 0.347 & 0.224 \\
 & TSen & 0.120 & 0.258 & 0.368 & 0.426 & 0.483 & 0.517 & 0.510 & 0.499 & 0.477 & 0.466 & 0.231 \\
 & TSenS & 0.115 & 0.234 & 0.325 & 0.388 & 0.450 & 0.545 & 0.510 & 0.496 & 0.488 & 0.470 & 0.211 \\
\midrule
\multirow{6}{*}{gte\_Qwen2-7B-instruct} & CSV & 0.072 & 0.134 & 0.196 & 0.234 & 0.297 & 0.330 & 0.311 & 0.301 & 0.291 & 0.290 & 0.217 \\
 & Mixed & 0.072 & 0.139 & 0.182 & 0.211 & 0.263 & 0.278 & 0.263 & 0.252 & 0.236 & 0.240 & 0.259 \\
 & HTML & 0.038 & 0.086 & 0.124 & 0.172 & 0.211 & 0.244 & 0.239 & 0.231 & 0.227 & 0.228 & 0.157 \\
 & Markdown & 0.077 & 0.134 & 0.196 & 0.249 & 0.278 & 0.282 & 0.280 & 0.279 & 0.282 & 0.282 & 0.271 \\
 & TSen & 0.096 & 0.163 & 0.225 & 0.311 & 0.349 & 0.354 & 0.328 & 0.330 & 0.341 & 0.329 & 0.270 \\
 & TSenS & 0.091 & 0.167 & 0.215 & 0.278 & 0.321 & 0.388 & 0.342 & 0.343 & 0.333 & 0.331 & 0.235 \\
\midrule
\multirow{6}{*}{jina-embeddings-v4} & CSV & 0.129 & 0.282 & 0.368 & 0.416 & 0.498 & 0.612 & 0.603 & 0.581 & 0.560 & 0.541 & 0.211 \\
 & Mixed & 0.163 & 0.306 & 0.426 & 0.483 & 0.541 & 0.641 & 0.634 & 0.609 & 0.573 & 0.547 & 0.254 \\
 & HTML & 0.158 & \textbf{0.368} & 0.474 & 0.584 & 0.646 & 0.689 & 0.687 & 0.667 & 0.644 & 0.634 & 0.229 \\
 & Markdown & 0.139 & 0.268 & 0.373 & 0.445 & 0.522 & 0.598 & 0.591 & 0.576 & 0.560 & 0.562 & 0.232 \\
 & TSen & 0.177 & 0.273 & 0.368 & 0.474 & 0.522 & 0.550 & 0.548 & 0.530 & 0.513 & 0.503 & \textbf{0.322} \\
 & TSenS & 0.158 & 0.239 & 0.364 & 0.459 & 0.522 & 0.560 & 0.560 & 0.544 & 0.533 & 0.521 & 0.282 \\
\midrule
\multirow{6}{*}{stella\_en\_1.5B\_v5} & CSV & 0.129 & 0.244 & 0.335 & 0.407 & 0.464 & 0.541 & 0.531 & 0.515 & 0.505 & 0.501 & 0.239 \\
 & Mixed & 0.096 & 0.172 & 0.244 & 0.287 & 0.340 & 0.526 & 0.476 & 0.448 & 0.417 & 0.405 & 0.182 \\
 & HTML & 0.048 & 0.105 & 0.172 & 0.220 & 0.258 & 0.349 & 0.330 & 0.309 & 0.297 & 0.292 & 0.137 \\
 & Markdown & 0.139 & 0.244 & 0.325 & 0.407 & 0.455 & 0.517 & 0.514 & 0.494 & 0.493 & 0.479 & 0.269 \\
 & TSen & \textbf{0.206} & 0.306 & 0.397 & 0.502 & 0.550 & 0.641 & 0.615 & 0.609 & 0.598 & 0.582 & 0.321 \\
 & TSenS & 0.163 & 0.292 & 0.397 & 0.493 & 0.560 & 0.632 & 0.632 & 0.617 & 0.602 & 0.588 & 0.258 \\
\bottomrule
\end{tabular}
}
\end{table*}

\subsection{Alternative Retrieval Paradigms}
\label{sec:appendix_alternative_retrieval}
Beyond standard dense embedding retrieval, we further evaluate several alternative retrieval paradigms, including table-specific retrievers (DTR) \citep{NQ-TABLES-DTR}, ColBERT-style late-interaction retrieval \citep{Colbert}, sparse lexical retrieval (BM25 and SPLADE \citep{SPLADE, SPLADE-v2,SPLADE-v3}), symbolic-neural hybrid retrieval, row-level multi-vector retrieval, cascaded retrieval (CRAFT) \citep{CRAFT}, and table-specific hybrid retrieval (THYME) \citep{THYME}.

\paragraph{Model Descriptions}
For DTR, we evaluate two variants: \texttt{tapas\_nq\_retriever\_large} and \texttt{tapas\_nq\_hn\_retriever\_large} \footnote{We use official DTR weights in a PyTorch reimplementation; other public adaptations (e.g., deepset/tapas-large-nq-reader) yielded inferior results.}. The ``nq'' suffix indicates that the model is fine-tuned on Natural Questions \citep{NQ-TABLES-DTR}, while ``hn'' denotes training with hard negative sampling. For ColBERT-style retrieval, we note that the original ColBERT architecture \citep{Colbert} is designed for passages and does not natively support 8K-token contexts. We therefore adopt three ColBERT-style models from HuggingFace that have been extended or adapted for longer sequences, namely \texttt{Reason-ModernColBERT}\footnote{\url{https://huggingface.co/lightonai/Reason-ModernColBERT}}, \texttt{SauerkrautLM-Multi-ModernColBERT}\footnote{\url{https://huggingface.co/VAGOsolutions/SauerkrautLM-Multi-ModernColBERT}}, and \texttt{reason-colBERT-150M-GTE-ModernColBERT}\footnote{\url{https://huggingface.co/fjmgAI/reason-colBERT-150M-GTE-ModernColBERT}}, as available open-source checkpoints without further fine-tuning. For sparse lexical retrieval, we consider both the classical term-matching approach BM25 and its learned counterpart \texttt{splade-v3} \citep{SPLADE-v3}\footnote{\url{https://huggingface.co/naver/splade-v3}}.

For row-level multi-vector retrieval, each table row (concatenated with its header) is independently embedded, and its similarity score with the query is computed separately; row-level scores are then aggregated into a single table-level score. We experimented with several aggregation strategies (mean, max, and top-$k$ averaging) and found max aggregation to perform best; we therefore report only the max-aggregation results. For hybrid retrieval, we perform a grid search over dense-BM25 interpolation weights using Qwen3-Embedding-4B as the dense retriever; the best-performing configuration uses a dense:BM25 ratio of 9:1.

We further reproduce CRAFT \citep{CRAFT}, a multi-model cascaded retrieval framework that combines sparse and dense retrieval with LLM-based text augmentation. To adapt CRAFT to our setting, we replace its original retrieval backbones with available open-source checkpoints while preserving the overall cascading procedure. Specifically, the retrieval cascade consists of \texttt{splade-v3}, Qwen3-Embedding-0.6B, and Qwen3-Embedding-4B. We evaluate two variants. \textit{CRAFT-Simplified} removes the LLM-based augmentation component and therefore consists solely of the retrieval cascade. The full CRAFT variants additionally apply LLM-based augmentation using either Qwen3-30B-A3B-2507-Thinking or Qwen3.6-27B. This setup allows us to separately assess the effectiveness of the cascaded retrieval architecture and the LLM-based augmentation component under our long-context table setting.

Finally, we reproduce THYME \citep{THYME}, a table-specific retrieval method that adopts a column-oriented representation and combines multiple retrieval signals over table fields. THYME is particularly relevant as a direct comparison with DTR: both are table-specific retrieval architectures trained specifically for table retrieval, rather than general-purpose retrievers applied to tables. We evaluate THYME using \texttt{bert-base-uncased}, \texttt{bert-large-uncased}, and \texttt{splade\_cocondenser\_selfdistil}, corresponding to the backbone configurations considered in the original work. We preserve its column-oriented retrieval paradigm as closely as possible without additional training on our benchmark.

\paragraph{Results}

Table~\ref{tab:alternative_retrieval_results} summarizes the retrieval performance across paradigms. Overall, most alternative retrieval paradigms underperform the strongest dense and row-level multi-vector baselines in our setting. DTR performs particularly poorly, with both variants achieving R@1 below 1\%. THYME performs better, with its strongest configuration reaching 0.062 R@1 and 0.371 DS@1, but remains substantially behind the strongest general-purpose retrieval methods.

CRAFT provides a competitive cascaded retrieval paradigm. \textit{CRAFT-Simplified}, which removes LLM-based augmentation, achieves 0.215 R@1 and 0.313 DS@1, essentially matching the dense-BM25 hybrid baseline. However, adding LLM-based augmentation does not provide additional gains: the full CRAFT variants using Qwen3-30B-A3B-2507-Thinking and Qwen3.6-27B achieve lower R@1 and DS@1 than the simplified variant. Notably, the Qwen3.6-27B configuration performs worse than the Qwen3-30B-A3B-2507-Thinking configuration, suggesting that using a later or more capable augmentation model does not necessarily improve retrieval under our setting. This is also consistent with the limited benefits of LLM-based augmentation observed across the augmentation strategies evaluated in Appendix~\ref{sec:appendix_augmentation}.

For THYME, we adopt three backbone model families as the base models for training, namely \texttt{bert-base-uncased}, \texttt{bert-large-uncased}, and \texttt{splade\_cocondenser\_selfdistil}. We follow the original training methodology as closely as possible to ensure a fair reproduction of the experimental setup. The resulting performance remains limited, with the \texttt{bert-large-uncased} configuration achieving the best R@1 of 0.062. Since THYME and DTR are both table-specific retrieval architectures trained specifically for table retrieval, their results provide a direct comparison with specialized table retrieval approaches. THYME substantially outperforms DTR but still falls behind general-purpose dense and row-level multi-vector retrieval. Under our long-context heterogeneous table setting, the specialized table retrieval architectures considered here therefore do not close the Target--Sibling gap relative to the stronger general-purpose baselines in this comparison.

\begin{table*}[htbp]
\centering
\small
\caption{Retrieval performance across alternative retrieval paradigms.}
\label{tab:alternative_retrieval_results}
\resizebox{\textwidth}{!}{
\begin{tabular}{l|l|ccc|c|c}
\toprule
\textbf{Model} & \textbf{Type} & \textbf{R@1} & \textbf{R@3} & \textbf{R@5} & \textbf{GR@1} & \textbf{DS@1} \\
\midrule
tapas\_nq\_hn\_retriever\_large & DTR & 0.010 & 0.014 & 0.019 & 0.038 & 0.250 \\
tapas\_nq\_retriever\_large & DTR & 0.005 & 0.010 & 0.010 & 0.005 & 1.000 \\
\midrule
Reason-ModernColBERT & ColBERT-style & 0.086 & 0.196 & 0.282 & 0.411 & 0.209 \\
SauerkrautLM-Multi-ModernColBERT & ColBERT-style & 0.172 & 0.474 & 0.632 & 0.766 & 0.225 \\
reason-colBERT-150M-GTE-ModernColBERT & ColBERT-style & 0.086 & 0.239 & 0.340 & 0.435 & 0.198 \\
\midrule
BM25 & Sparse & 0.005 & 0.005 & 0.014 & 0.005 & 1.000 \\
splade-v3 & Sparse & 0.091 & 0.244 & 0.368 & 0.392 & 0.232 \\
Qwen3-E4B + BM25 hybrid & Hybrid & 0.215 & 0.455 & 0.612 & 0.689 & 0.312 \\
\midrule
Qwen3-E0.6B & Row-level multi-vector & 0.263 & 0.488 & 0.627 & 0.699 & 0.377 \\
Qwen3-E4B & Row-level multi-vector & 0.249 & 0.507 & 0.679 & 0.732 & 0.340
\\
Qwen3-E8B & Row-level multi-vector & 0.282 & 0.512 & 0.660 & 0.694 & 0.407
\\
\midrule
\textit{CRAFT-Simplified} & Cascade & 0.215 & 0.464 & 0.646 & 0.689 & 0.313 \\
CRAFT (Qwen3-30B-A3B-2507-Thinking) & Cascade + LLM Aug. & 0.196 & 0.450 & 0.589 & 0.670 & 0.293 \\
CRAFT (Qwen3.6-27B) & Cascade + LLM Aug. & 0.187 & 0.455 & 0.593 & 0.684 & 0.273 \\
\midrule
bert-base-uncased & THYME & 0.014 & 0.062 & 0.086 & 0.067 & 0.214 \\
bert-large-uncased & THYME & 0.062 & 0.096 & 0.166 & 0.168 & 0.371 \\
splade\_cocondenser\_selfdistil & THYME & 0.014 & 0.096 & 0.158 & 0.134 & 0.107 \\
\bottomrule
\end{tabular}
}
\end{table*}

\paragraph{Analysis}

Across the paradigms we evaluate, the results reveal a consistent Semantic-Answerability Gap between retrieving a relevant table group and identifying the uniquely answerable table. DTR and THYME, despite being table-specific architectures trained for table retrieval, achieve substantially lower R@1 than the strongest general-purpose retrievers in this comparison. ColBERT-style and sparse retrievers improve recall considerably, but their DS@1 remains limited. The dense-BM25 hybrid and CRAFT-Simplified provide competitive performance, yet neither substantially closes the gap between group-level retrieval and exact table retrieval.

Row-level multi-vector retrieval is the strongest first-stage paradigm in this comparison, achieving the highest R@1 and GR@1 among the evaluated approaches. Nevertheless, its DS@1 remains substantially below its GR@1, indicating that even strong retrievers in this set can reach the correct topical group without reliably selecting the table that contains sufficient evidence for answering the query. This pattern persists across the dense, late-interaction, sparse, hybrid, cascaded, and table-specific paradigms we test.

Within our evaluation, the observed limitation is therefore not explained by the choice of first-stage architecture alone. Distinguishing the uniquely answerable table appears to require fine-grained matching of schema, values, and query constraints that the semantic, lexical, and table-specific representations we evaluate do not capture adequately. This is consistent with answerability acting as a retrieval signal that is only partially reflected by relevance-oriented similarity.

\subsection{LLM-Augmented Dense Retrieval}
\label{sec:appendix_augmentation}

We further evaluate several augmentation strategies commonly used in retrieval pipelines to determine whether the observed Semantic-Answerability Gap can be mitigated by adding auxiliary semantic signals.

\paragraph{Experimental Setup}

Inspired by prior work~\citep{HyDE, LameR}, we evaluate two semantic augmentation strategies using Qwen3-30B-A3B-Thinking-2507:

\begin{itemize}

\item \textbf{Meta-Info Generation.}  
For each table, we generate descriptive metadata summarizing table themes, column semantics, and value distributions. 
The generated metadata is concatenated with the table content before embedding.

\item \textbf{Query-to-Table (Q2T).}  
Given a query, the model first generates a hypothetical ``answer table''. 
Retrieval is then performed by matching this generated table against candidate tables in the corpus.

\end{itemize}

\paragraph{Results}

Table~\ref{tab:appendix_aug_results_full} summarizes the results under the Qwen3-Embedding-8B retrieval backbone.

\begin{table*}[htbp]
\centering
\small
\caption{Retrieval performance of augmentation strategies across different embedding models.}
\label{tab:appendix_aug_results_full}
\resizebox{\textwidth}{!}{
\begin{tabular}{l|l|ccc|ccc|c}
\toprule
\textbf{Method} & \textbf{Retrieval Model} & \textbf{R@1} & \textbf{R@3} & \textbf{R@5} & \textbf{GR@1} & \textbf{GR@3} & \textbf{GR@5} & \textbf{DS@1} \\
\midrule
\multirow{3}{*}{Meta-Augmentation}
 & Qwen3-Embedding-0.6B & 0.144 & 0.426 & 0.584 & 0.646 & 0.770 & 0.833 & 0.222 \\
 & Qwen3-Embedding-4B   & 0.196 & 0.512 & 0.670 & 0.746 & 0.833 & 0.880 & 0.263 \\
 & Qwen3-Embedding-8B   & 0.153 & 0.498 & 0.660 & 0.742 & 0.785 & 0.837 & 0.206 \\
\midrule
\multirow{3}{*}{Query-to-Table}
 & Qwen3-Embedding-0.6B & 0.153 & 0.388 & 0.478 & 0.550 & 0.651 & 0.679 & 0.278 \\
 & Qwen3-Embedding-4B   & 0.144 & 0.368 & 0.545 & 0.541 & 0.660 & 0.713 & 0.265 \\
 & Qwen3-Embedding-8B   & 0.153 & 0.402 & 0.502 & 0.565 & 0.632 & 0.684 & 0.271 \\
\bottomrule
\end{tabular}
}
\end{table*}

\paragraph{Analysis}

Both semantic augmentation methods provide only limited improvements over the baseline dense retrievers.
Meta-information generation slightly improves recall metrics but does not substantially improve the ranking of Target Tables.
The Query-to-Table strategy frequently introduces hallucinated structures in the generated tables, which inject additional noise into the retrieval process.

These observations suggest that, among the strategies we evaluate, simply enriching the semantic representation of tables is not sufficient to resolve ambiguity among structurally similar tables on \ours{}.
In contrast, Answerability-Aware Reranking (AAR) is more effective in our setting because it enables direct query-table interaction during scoring.

\subsection{Retrieval and Downstream QA Performance by Query Type}
\label{app:retrieval_qa_by_type}

To further characterize where retrieval and reasoning difficulties arise, we report a breakdown of retrieval and downstream QA performance by query type using Qwen3-Embedding-8B as the retriever.

\begin{table*}[t]
\centering
\caption{Retrieval performance by query type using Qwen3-Embedding-8B.}
\label{tab:app_retrieval_by_type}
\begin{tabular}{l|ccc|c|c}
\toprule
Type & R@1 & R@3 & R@5 & GR@1 & DS@1 \\
\midrule
All & 0.182 & 0.469 & 0.665 & 0.670 & 0.271 \\
EM  & 0.224 & 0.531 & 0.724 & 0.755 & 0.297 \\
SF  & 0.182 & 0.418 & 0.600 & 0.564 & 0.323 \\
SA  & 0.107 & 0.411 & 0.625 & 0.625 & 0.171 \\
\bottomrule
\end{tabular}
\end{table*}

\begin{table*}[t]
\centering
\caption{Downstream QA performance by query type using Qwen3-Embedding-8B.}
\label{tab:app_qa_by_type}
\begin{tabular}{l|c|ccc|ccc}
\toprule
Type & Oracle & QA@1 & QA@3 & QA@5 & Eff@1 & Eff@3 & Eff@5 \\
\midrule
All & 0.755 & 0.139 & 0.266 & 0.330 & 1.014 & 0.751 & 0.657 \\
EM  & 0.884 & 0.194 & 0.415 & 0.486 & 0.977 & 0.885 & 0.759 \\
SF  & 0.711 & 0.129 & 0.180 & 0.204 & 0.998 & 0.605 & 0.479 \\
SA  & 0.571 & 0.054 & 0.089 & 0.179 & 0.876 & 0.381 & 0.500 \\
\bottomrule
\end{tabular}
\end{table*}

The breakdown reveals three main patterns. First, EM is the easiest query type overall, achieving the highest group recall, exact retrieval, and downstream QA performance among the three types. Even so, its DS@1 remains near the random-selection baseline, indicating that a Semantic-Answerability Gap is present even for the easiest queries. Second, SA is the hardest type for exact answerability discrimination, obtaining the lowest R@1 and DS@1 of the three types. Third, coarse retrieval and within-group discrimination behave as distinct sources of difficulty: SF has the lowest GR@1 among the three types, yet its DS@1 is higher than that of both EM and SA, showing that difficulty in locating the correct semantic group is not identical to difficulty in identifying the target within that group once it has been located.

These results suggest that retrieval difficulty and reasoning difficulty are related but not interchangeable. Given SA's low retrieval performance and low Oracle score, the non-monotonic Eff@k pattern for this type should be interpreted with caution: SA has comparatively few queries, so the effective sample after retrieval is small and downstream QA estimates can be unstable. We therefore refrain from drawing strong conclusions from minor numeric variations in Eff@k.

\subsection{Effect of Table Length on Retrieval Performance}
\label{sec:appendix_length_scaling}

\begin{figure}[t]\centering\includegraphics[width=0.95\textwidth]{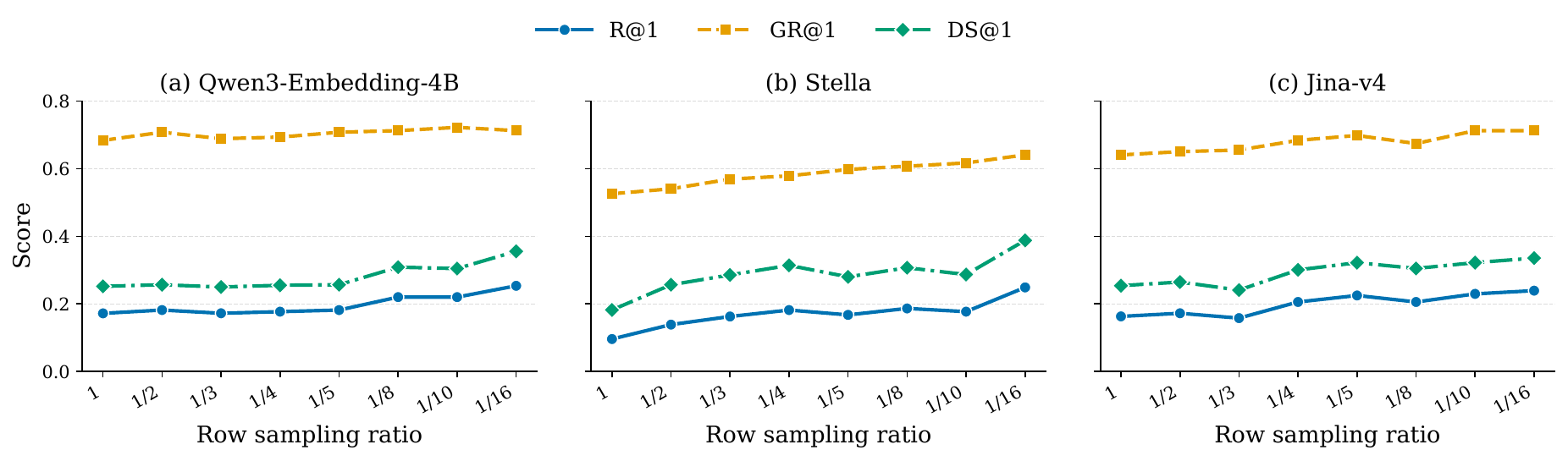} \caption{Effect of row-sampling ratio on retrieval performance. Each panel corresponds to one embedding model (Qwen3-E4B, Stella, Jina); curves show $R@1$, $GR@1$, and $DS@1$ against the retained-row ratio $1/n$.} \label{fig:length_scaling}\end{figure}

\paragraph{Setup.}
On \ours{}, the evaluated dense retrievers have difficulty locating the uniquely answerable Target among structurally similar Sibling Distractor Tables. To test whether this difficulty stems from semantic dilution caused by excessively long tables, we apply row sampling to every table in the candidate pool, reducing each table to $1/n$ of its original number of rows, where $n \in \{2, 3, 4, 5, 8, 10, 16\}$. For the Target Table, all rows relevant to the answer are retained, so that answer-bearing evidence is not removed during sampling. We report R@1, GR@1, and DS@1 across sampling ratios on three embedding models (Qwen3-E4B, Stella, and Jina); results are shown in Figure~\ref{fig:length_scaling}.

\paragraph{Results and analysis.}
All three models exhibit a consistent trend: as the sampling ratio decreases (i.e., tables become smaller), R@1, GR@1, and DS@1 improve to varying degrees. For Qwen3-E4B, R@1 rises from 0.172 at full length to 0.254 in the $1/16$ regime, with GR@1 improving from 0.684 to approximately 0.713. The improvement is more pronounced for Stella and Jina; for instance, Stella's DS@1 rises from 0.182 at full length to 0.388 under $1/16$ sampling, while Jina's DS@1 rises from 0.254 to 0.336. This is consistent with the hypothesis that excessive table length dilutes the query-relevant signal for these dense embeddings.

However, even at the lowest sampling ratio (i.e., the smallest table size), the absolute values of R@1 and DS@1 remain low: none of the three models exceeds an R@1 of 0.26 or a DS@1 of 0.39. Shrinking tables therefore partially mitigates retrieval difficulty for these models, but does not eliminate Target--Sibling confusion. Even when sibling tables are compressed to a small scale, the evaluated models still do not consistently identify the uniquely answerable target. Within this length-scaling experiment, residual errors are thus unlikely to be explained by table length alone and are more consistent with limited sensitivity to fine-grained structural or value-level distinctions among near-duplicate tables.

\section{Real-TCR: Natural-Negative Validation}
\label{app:real_tcr}

TCR-Bench is intentionally constructed as a controlled stress test in which sibling tables share coarse semantic and structural characteristics while differing in whether the evidence required by a query is present. While this controlled construction makes answerability measurable and enables systematic analysis, it also raises a natural concern: whether the observed Semantic-Answerability Gap is specific to the benchmark construction or also occurs in naturally organized table collections.

To address this concern, we construct \textbf{Real-TCR}, an additional evaluation collection from naturally occurring tables in a real-world environment. Unlike TCR-Bench, its candidate tables are not generated through answerability-targeted transformations. Instead, they follow natural partitions already present in the underlying data, such as by time window or organizational unit. As a result, the tables exhibit naturally occurring variation in schema, column sets, and length. The data are substantially different in both content and organization from the open-source data used in TCR-Bench, providing an additional evaluation in a naturally organized setting.

Real-TCR contains 77 queries and 217 tables, while Real-TCR-small contains 38 queries and 117 tables, with candidate groups capped at 10 tables. Table lengths range from 3K to 8K tokens. The corresponding random DS@1 baselines are 0.047 and 0.100, respectively.

\begin{table*}[t]
\centering
\caption{Statistics of the naturally occurring Real-TCR collections.}
\label{tab:real_tcr_statistics}
\begin{tabular}{lcccc}
\toprule
Benchmark & \#Queries & \#Tables & Table Length & Random DS@1 \\
\midrule
Real-TCR & 77 & 217 & 3K--8K & 0.047 \\
Real-TCR-small & 38 & 117 & 3K--8K & 0.100 \\
\bottomrule
\end{tabular}
\end{table*}

The queries and target tables were manually annotated by personnel of an industry partner organization, followed by constraint verification, false-negative retrieval, and manual inspection. Additional information was available during annotation to perform global consistency checks and verify target uniqueness. The underlying data consist of heterogeneous real-world operational records obtained under a data-use agreement. All data were de-identified prior to use and cannot be publicly released due to contractual constraints; Real-TCR therefore serves as a non-releasable external validation rather than a publicly reproducible benchmark.

\begin{table*}[t]
\centering
\caption{Retrieval results on naturally occurring Real-TCR collections.}
\label{tab:real_tcr_natural}
\begin{tabular}{l|ccccc}
\toprule
Benchmark & Model & R@1 & GR@1 & DS@1 & Random DS@1 \\
\midrule
\multirow{3}{*}{Real-TCR}  & Qwen3-E4B & 0.039 & 0.909 & 0.043 & 0.047 \\
& Jina & 0.143 & 0.948 & 0.151 & 0.047 \\
& Stella & 0.104 & 0.870 & 0.119 & 0.047 \\
\midrule
\multirow{3}{*}{Real-TCR-small}  & Qwen3-E4B & 0.079 & 0.921 & 0.086 & 0.100 \\
& Jina & 0.184 & 0.947 & 0.194 & 0.100 \\
& Stella & 0.105 & 0.868 & 0.121 & 0.100 \\
\bottomrule
\end{tabular}
\end{table*}

The same qualitative pattern observed on TCR-Bench appears for the evaluated retrievers on this natural collection. For example, Jina achieves GR@1 = 0.948 on Real-TCR but only R@1 = 0.143 and DS@1 = 0.151. Qwen3-E4B achieves GR@1 = 0.909 but DS@1 = 0.043. Thus, these retrievers can identify the relevant semantic group with high accuracy while still struggling to select the exact table containing sufficient information to answer the query.

Real-TCR therefore provides a natural-negative validation of the phenomenon under a substantially different data organization. Unlike TCR-Bench, where distractors are constructed through controlled perturbations, the candidate tables in Real-TCR arise from naturally occurring partitions of real-world data. The persistence of the same qualitative pattern on this collection provides additional evidence that the observed gap is not an artifact unique to the construction procedure of \ours{}. We do not claim that every real-world data collection exhibits the same quantitative difficulty; rather, the shared qualitative pattern across the controlled benchmark and this naturally organized collection supports the diagnostic relevance of the finding beyond synthetic sibling construction alone.

\section{Complete Surface Variation Results}
\label{app:surface_variation_results}

This section reports the query-variant definitions, complete retrieval numbers, and pairwise consistency statistics for the surface-variation analysis in Section~\ref{sec:surface_variations}. The results support the finding that Target--Sibling discrimination is not primarily an artifact of serialization format or query wording in our evaluation; they do not introduce a separate conclusion.

\subsection{Query Variant Definitions}
\label{app:query-variants}

We evaluate three query variants, as summarized in Table~\ref{tab:query-variants}.

\begin{table*}[htbp]
\centering
\caption{Summary of the three query variants.}
\label{tab:query-variants}
\begin{tabular}{p{2.2cm}|p{5cm}|p{5cm}}
\hline
\textbf{Variant} & \textbf{Format} & \textbf{Example} \\
\hline
Final NL & Natural-language question & “Which country won more than 29 gold medals in the 2016 Olympics?” \\
\hline
Sentential & Conditions expressed as separate sentences; target requested at the end & “The year is 2016. The number\_of\_gold\_medals is greater than 29. Country is?” \\
\hline
Template & Structured templates with explicit placeholders for attributes, values, and conditions & “What is the Country when Year = 2016 AND Gold > 29?” \\
\hline
\end{tabular}
\end{table*}

For template queries, we use three sets of fixed patterns, each targeting a different query intention:

\begin{itemize}
    \item \textbf{EM (Exact-Match Lookup):} Retrieve target values for rows satisfying conditions.  
    \emph{Patterns:} \texttt{"What is the \{target\} when \{conds\}?"}, \texttt{"Please find the \{target\} for rows where \{conds\}."}, \texttt{"List the \{target\} for the entries with \{conds\}."}, \texttt{"Which \{target\} values correspond to \{conds\}?"}

    \item \textbf{SF (Simple Filtering):} Similar to EM, with consistent emphasis on row-based filtering.  
    \emph{Patterns:} \texttt{"What is the \{target\} when \{conds\}?"}, \texttt{"List the \{target\} for rows where \{conds\}."}, \texttt{"Which \{target\} values correspond to rows where \{conds\}?"}, \texttt{"Please find the \{target\} for entries with \{conds\}."}

    \item \textbf{SA (Simple Aggregation):} Apply aggregation functions (e.g., COUNT, SUM, AVG) over the target.  
    \emph{Patterns:} \texttt{"What is the \{agg\} of \{target\} when \{conds\}?"}, \texttt{"Compute the \{agg\} of \{target\} for rows where \{conds\}."}, \texttt{"Find the \{agg\} value of \{target\} corresponding to rows where \{conds\}."}, \texttt{"Please calculate the \{agg\} of \{target\} for entries with \{conds\}."}
\end{itemize}

The condition placeholder \texttt{\{conds\}} is instantiated as conjunctions of attribute–value comparisons using symbolic operators (e.g., \texttt{Year = 2016 AND Gold > 29}).

\begin{table}[htbp]
\centering
\caption{Complete retrieval results on the Diagnostic Subset under different table serialization formats.}
\label{tab:serialization_formats}
\begin{tabular}{l|l|ccc|c|c}
\toprule
\multirow{2}{*}{\textbf{Model}} & \multirow{2}{*}{\textbf{Format}} & \multicolumn{3}{c|}{\textbf{R@k}} & \multirow{2}{*}{\textbf{GR@1}} & \multirow{2}{*}{\textbf{DS@1}} \\
\cmidrule(lr){3-5}
 & & \textbf{@1} & \textbf{@3} & \textbf{@5} & & \\ 
\midrule
\multirow{6}{*}{Qwen3-E4B}
 & CSV & 0.186 & 0.571 & \textbf{0.771} & 0.814 & 0.228 \\
 & Mixed & 0.186 & 0.586 & 0.729 & 0.814 & 0.228 \\
 & HTML & 0.200 & 0.543 & 0.757 & \textbf{0.857} & 0.233 \\
 & Markdown & \textbf{0.257} & 0.557 & 0.729 & 0.814 & 0.316 \\
 & TSen & 0.200 & 0.529 & 0.743 & 0.800 & 0.250 \\
 & TSenS & 0.243 & \textbf{0.614} & 0.714 & 0.757 & \textbf{0.321} \\
\midrule
\multirow{6}{*}{Stella}
 & CSV & 0.171 & 0.414 & 0.543 & 0.671 & 0.255 \\
 & Mixed & 0.143 & 0.357 & 0.500 & 0.657 & 0.217 \\
 & HTML & 0.086 & 0.257 & 0.414 & 0.629 & 0.136 \\
 & Markdown & 0.171 & 0.386 & 0.514 & 0.629 & 0.273 \\
 & TSen & \textbf{0.314} & \textbf{0.500} & 0.643 & \textbf{0.786} & \textbf{0.400} \\
 & TSenS & 0.286 & 0.471 & \textbf{0.671} & 0.771 & 0.370 \\
\bottomrule
\end{tabular}
\end{table}

\begin{table}[htbp]
\centering
\small
\caption{Complete retrieval results under different query formulations.}
\label{tab:query_diag}
\begin{tabular}{l|l|ccc|c|c}
\toprule
\multirow{2}{*}{\textbf{Model}} & \multirow{2}{*}{\textbf{Query Type}} & \multicolumn{3}{c|}{\textbf{R@k}} & \multirow{2}{*}{\textbf{GR@1}} & \multirow{2}{*}{\textbf{DS@1}} \\ 
\cmidrule(lr){3-5}
 & & \textbf{@1} & \textbf{@3} & \textbf{@5} & & \\ 
\midrule
\multirow{3}{*}{Qwen3-E4B}
 & Final (NL) & 0.186 & \textbf{0.586} & \textbf{0.729} & \textbf{0.814} & 0.228 \\
 & Sentential & 0.229 & 0.471 & \textbf{0.729} & 0.800 & 0.286 \\
 & Template & \textbf{0.257} & 0.571 & 0.714 & 0.786 & \textbf{0.327} \\ 
\midrule
\multirow{3}{*}{Stella}
 & Final (NL) & \textbf{0.143} & \textbf{0.357} & \textbf{0.500} & 0.657 & \textbf{0.217} \\
 & Sentential & 0.100 & 0.343 & \textbf{0.500} & 0.629 & 0.159 \\
 & Template & 0.100 & 0.329 & 0.471 & \textbf{0.714} & 0.140 \\
\bottomrule
\end{tabular}
\end{table}

\subsection{Consistency Analysis Across Surface Variations}
\label{appendix:consistency}

We further compare retrieval outputs pairwise across serialization formats and query formulations, using the metrics below.

\begin{itemize}
    \item \textbf{Hit Consistency}: the proportion of instances where top-1 retrieval outcomes (hit vs.\ miss) are identical across two configurations.
    
    \item \textbf{Partial Consistency}: the proportion of cases where the top-1 table from the base configuration appears within the top-5 results of the comparison configuration.
    
    \item \textbf{Ranking Consistency}: Rank-Biased Overlap (RBO, $p=0.9$)~\citep{RBO} between Top-5 ranking lists. RBO emphasizes higher-ranked items while accounting for list overlap.
    
    \item \textbf{Top-1 Identity}: the percentage of cases where the exact same table ID appears at rank 1 in both configurations.
\end{itemize}

High Hit or Partial Consistency with lower Ranking Consistency or Top-1 Identity corresponds to stable hit/miss behavior and overlapping candidates, but shifting order within the list.

\subsection{Serialization Format Consistency}
\label{appendix:serialization_consistency}

Table~\ref{tab:format_consistency_extended_appendix} reports consistency metrics across serialization formats using the Mixed format as the base configuration.

\begin{table*}[tp]
\centering
\small
\caption{Consistency analysis across serialization formats and narrative structures.}
\label{tab:format_consistency_extended_appendix}
\begin{tabular}{l|l|cccc}
\toprule
\textbf{Model} & \textbf{Config 1 vs. 2} & \textbf{Hit Consis.} & \textbf{Ranking (RBO)} & \textbf{Partial Consis.} & \textbf{Top-1 Id.} \\ \midrule
\multirow{6}{*}{Qwen3-E4B} 
& Mixed vs. CSV & 85.71\% & 29.22\% & 97.14\% & 58.57\% \\
& Mixed vs. HTML & 87.14\% & 27.76\% & 94.29\% & 45.71\% \\
& Mixed vs. Markdown & 81.43\% & 28.47\% & 97.14\% & 48.57\% \\
& Mixed vs. TSen & 75.71\% & 25.88\% & 92.86\% & 38.57\% \\
& Mixed vs. TSenS & 82.86\% & 25.06\% & 88.57\% & 41.43\% \\
& \textbf{TSen vs. TSenS} & \textbf{84.29\%} & \textbf{30.14\%} & \textbf{94.29\%} & \textbf{58.57\%} \\ \midrule
\multirow{6}{*}{Stella} 
& Mixed vs. CSV & 88.57\% & 24.55\% & 88.57\% & 52.86\% \\
& Mixed vs. HTML & 82.86\% & 19.98\% & 71.43\% & 42.86\% \\
& Mixed vs. Markdown & 88.57\% & 25.87\% & 92.86\% & 55.71\% \\
& Mixed vs. TSen & 77.14\% & 20.43\% & 82.86\% & 34.29\% \\
& Mixed vs. TSenS & 74.29\% & 20.05\% & 78.57\% & 35.71\% \\
& \textbf{TSen vs. TSenS} & \textbf{91.43\%} & \textbf{34.43\%} & \textbf{95.71\%} & \textbf{77.14\%} \\
\bottomrule
\end{tabular}
\end{table*}

Hit Consistency and Partial Consistency are high for most format pairs, so target hit/miss and the base top-1's presence in the comparison top-5 are largely preserved. The TSen vs.\ TSenS comparison is also high, which is consistent with limited sensitivity to within-row attribute order. Ranking Consistency is lower (RBO around 20--30\%): candidate sets remain similar, but order within the top-5 often changes. These statistics are consistent with coarse semantic matching remaining stable under serialization changes, while fine-grained scores still move.

\subsection{Query Formulation Consistency}
\label{appendix:query_consistency}

Table~\ref{tab:query_consistency_appendix} reports the same metrics using the Final (NL) query as the base configuration.

\begin{table*}[tp]
\centering
\small
\caption{Consistency analysis across query variations (Base: Final Query).}
\label{tab:query_consistency_appendix}
\begin{tabular}{l|l|cccc}
\toprule
\textbf{Model} & \textbf{vs. Query} & \textbf{Hit Consis.} & \textbf{Ranking (RBO)} & \textbf{Partial Consis.} & \textbf{Top-1 Id.} \\ \midrule
\multirow{2}{*}{Qwen3-E4B} 
& Template & 84.29\% & 31.67\% & 97.14\% & 58.57\% \\
& Sentential & 84.29\% & 27.65\% & 90.00\% & 48.57\% \\ \midrule
\multirow{2}{*}{Stella} 
& Template & 95.71\% & 30.92\% & 94.29\% & 70.00\% \\
& Sentential & 90.00\% & 28.46\% & 90.00\% & 60.00\% \\
\bottomrule
\end{tabular}
\end{table*}

Hit and Partial Consistency remain high under query reformulation, while Top-1 Identity is only moderate. As with serialization, this indicates that candidate membership is largely preserved and that remaining differences are mainly ranking shifts. Together with Table~\ref{tab:query_diag}, the results are consistent with the observation that query phrasing is not the primary driver of the Target--Sibling gap for the two evaluated models.

\section{Detailed Dense Retrieval Diagnostics}
\label{app:dense_retrieval_diagnostics}

This section provides detailed perturbation settings, illustrative examples, and complete results for the diagnostic experiments in Section~\ref{sec:retrieval_behavior}.

\subsection{Semantic-volume perturbation}
We investigate whether dense retrieval is sensitive to the amount of semantic content or to the information required for identification. For each table row, we construct three sentence-based representations. \textbf{Sen} converts the complete row into a declarative sentence while preserving the original attribute order. \textbf{SenS} applies the same conversion but randomly shuffles the attribute order within the sentence. \textbf{SenSS} retains only a subset of column-value pairs that is sufficient to uniquely identify the relevant row.

For illustration, given a row containing the attributes \textit{Year}, \textit{Country}, \textit{Product}, and \textit{Sales}, the representations can be written as follows:

\begin{quote}
\textbf{Sen:} In 2020, the country was France, the product was Wine, and sales were 120.

\textbf{SenS:} The product was Wine, sales were 120, in 2020, and the country was France.

\textbf{SenSS:} In 2020, the product was Wine and sales were 120.
\end{quote}

Sen and SenS preserve the complete row, whereas SenSS retains only a subset sufficient for identification in the corresponding diagnostic case. SenS isolates sensitivity to attribute order, while SenSS examines whether retrieval benefits from additional semantic content beyond what is logically sufficient for identification.

We evaluate these representations at both the inter-table and intra-table levels. At the inter-table level, the task is to identify the Target Table from the candidate tables. At the intra-table level, the task is to retrieve the relevant row or a chunk containing it from the Target Table.

The results show that Sen and SenS generally outperform SenSS despite SenSS retaining sufficient information for identification, while the difference between Sen and SenS remains small. This indicates that the evaluated embeddings are more responsive to the amount of semantic content than to attribute order or minimal information sufficiency.

At the intra-table level, row-level retrieval substantially outperforms retrieval over 10-line chunks. Figure~\ref{fig:sentential_intra_full} presents the corresponding results. This contrast indicates that the evaluated dense embeddings can capture local row-level content while suffering from information dilution at larger granularities.

\subsubsection{Detailed Intra-table Retrieval Results}
\label{app:sentential_intra_table}
Table~\ref{tab:sentential_intra_full} reports the exact R@k values for row-level and chunk-level retrieval under the Sen, SenS, and SenSS representations.

\begin{table}[htbp]
\centering
\small
\caption{Intra-table retrieval by granularity and representation: row-level retrieval outperforms 10-line chunks for both evaluated models; SenSS yields lower accuracy despite retaining a logically sufficient subset.}
\label{tab:sentential_intra_full}
\begin{tabular}{l|l|l|ccc}
\toprule
\textbf{Model} & \textbf{Config} & \textbf{Type} & \textbf{R@1} & \textbf{R@3} & \textbf{R@5} \\
\midrule

\multirow{6}{*}{Qwen3-E4B}
& \multirow{3}{*}{Row}
& Sen & \textbf{0.871} & 0.903 & \textbf{0.929} \\
& & SenS & 0.845 & \textbf{0.916} & \textbf{0.929} \\
& & SenSS & 0.619 & 0.735 & 0.800 \\
\cmidrule{2-6}

& \multirow{3}{*}{Chunk}
& Sen & \textbf{0.387} & \textbf{0.587} & \textbf{0.742} \\
& & SenS & 0.335 & 0.561 & 0.723 \\
& & SenSS & 0.284 & 0.490 & 0.665 \\

\midrule

\multirow{6}{*}{Stella}
& \multirow{3}{*}{Row}
& Sen & \textbf{0.852} & \textbf{0.890} & \textbf{0.903} \\
& & SenS & 0.800 & 0.858 & 0.877 \\
& & SenSS & 0.619 & 0.729 & 0.755 \\
\cmidrule{2-6}

& \multirow{3}{*}{Chunk}
& Sen & \textbf{0.497} & \textbf{0.710} & \textbf{0.832} \\
& & SenS & 0.471 & 0.677 & 0.819 \\
& & SenSS & 0.400 & 0.645 & 0.768 \\

\bottomrule
\end{tabular}
\end{table}

\subsubsection{Comparison of 10-line and 20-line Chunk Retrieval}
\label{app:10-20-line}
Table~\ref{tab:chunk_comparison} further compares 10-line and 20-line chunk settings. Increasing chunk size does not yield consistent improvements: for Qwen3-E4B, $R@1$ rises from $0.387$ (10-line) to $0.432$ (20-line) under Sen, and SenS improves by a similar margin, whereas SenSS declines; Stella shows modest $R@1$ gains, but SenSS $R@3$ drops from $0.645$ to $0.561$. Aggregating more rows therefore does not reliably help and can further dilute query-relevant information.

\begin{table*}[htbp]
\centering
\small
\caption{10-line vs 20-line Chunk Retrieval Comparison.}
\label{tab:chunk_comparison}
\begin{tabular}{l|l|ccc|ccc}
\toprule
\textbf{Model} & \textbf{Type} & \textbf{10-line R@1} & \textbf{20-line R@1} & \textbf{$\Delta$} & \textbf{10-line R@3} & \textbf{20-line R@3} & \textbf{$\Delta$} \\
\midrule
\multirow{3}{*}{Qwen3-E4B}
 & Sen & 0.387 & 0.432 & +0.045 & 0.587 & 0.619 & +0.032 \\
 & SenS & 0.335 & 0.387 & +0.052 & 0.561 & 0.619 & +0.058 \\
 & SenSS & 0.284 & 0.271 & -0.013 & 0.490 & 0.445 & -0.045 \\
\midrule
\multirow{3}{*}{Stella}
 & Sen & 0.497 & 0.516 & +0.019 & 0.710 & 0.742 & +0.032 \\
 & SenS & 0.471 & 0.516 & +0.045 & 0.677 & 0.703 & +0.026 \\
 & SenSS & 0.400 & 0.413 & +0.013 & 0.645 & 0.561 & -0.084 \\
\bottomrule
\end{tabular}
\end{table*}

\subsection{Row-column binding perturbation}

We next examine whether dense representations preserve \textbf{row-column bindings}, namely, whether values remain associated with the columns and rows in which they originally occur. Analogous to the three sentence-based representations above, for each table in the Diagnostic Subset we construct a Minimal Contrastive Set with three configurations: \textbf{Ori} is the original table; \textbf{RID} augments each cell with its row identifier; and \textbf{RID+Shuf} further shuffles values within each column while keeping the row identifiers attached to their original values. Thus Ori and RID preserve row-column bindings, with RID making row identity explicit, whereas RID+Shuf retains the same values, columns, row identifiers, and entity alignments, but breaks the original row-column bindings. In other words, all three variants preserve the alignment between entities and their associated values, allowing us to isolate the effect of row-level bindings.

For example, given an original table
\[
\renewcommand{\arraystretch}{1.15}
\setlength{\tabcolsep}{8pt}
\begin{array}{|c|c|}
\hline
\mathbf{A} & \mathbf{B}\\
\hline
1 & 2\\
\hline
3 & 4\\
\hline
\end{array}
\]
the \textbf{RID} configuration augments each value with its original row identifier:
\[
\begin{array}{|c|c|}
\hline
\mathbf{A} & \mathbf{B}\\
\hline
\mathrm{r0{:}1} & \mathrm{r0{:}2}\\
\hline
\mathrm{r1{:}3} & \mathrm{r1{:}4}\\
\hline
\end{array}
\]
The \textbf{RID+Shuf} configuration then independently shuffles values within each column while retaining their row identifiers:
\[
\begin{array}{|c|c|}
\hline
\mathbf{A} & \mathbf{B}\\
\hline
\mathrm{r0{:}1} & \mathrm{r1{:}4}\\
\hline
\mathrm{r1{:}3} & \mathrm{r0{:}2}\\
\hline
\end{array}
\]

Full results across all formats appear in Table~\ref{tab:shuffle_full}. This controlled construction tests whether the embedding is sensitive to fine-grained row--column associations rather than to shared lexical content alone: RID+Shuf keeps explicit row indices and the same cell values, yet rearranges their within-column placements.

\begin{table*}[htbp]
\centering
\small
\caption{Full retrieval performance across all table formats and configurations.}
\label{tab:shuffle_full}
\begin{tabular}{l|l|l|ccc|c|c}
\toprule
Model & Config & Format & R@1 & R@2 & R@3 & GR@1 & DS@1 \\
\midrule

\multirow{4}{*}{Qwen3-E4B}
 & Ori & CSV & 0.400 & 0.671 & 0.714 & 0.686 & 0.583 \\
 & Ori & Mixed & 0.443 & 0.700 & 0.800 & 0.671 & 0.660 \\
 & Ori & HTML & 0.386 & 0.686 & 0.800 & 0.671 & 0.574 \\
 & Ori & Markdown & 0.343 & 0.671 & 0.729 & 0.643 & 0.533 \\

\midrule

\multirow{4}{*}{Stella}
 & Ori & CSV & 0.229 & 0.529 & 0.571 & 0.557 & 0.410 \\
 & Ori & Mixed & 0.229 & 0.443 & 0.529 & 0.471 & 0.485 \\
 & Ori & HTML & 0.257 & 0.414 & 0.514 & 0.443 & 0.581 \\
 & Ori & Markdown & 0.214 & 0.514 & 0.586 & 0.500 & 0.429 \\

\midrule

\multirow{4}{*}{Qwen3-E4B}
 & RID & CSV & 0.357 & 0.729 & 0.757 & 0.629 & 0.568 \\
 & RID & Mixed & 0.371 & 0.700 & 0.743 & 0.671 & 0.553 \\
 & RID & HTML & 0.314 & 0.700 & 0.786 & 0.686 & 0.458 \\
 & RID & Markdown & 0.400 & 0.657 & 0.743 & 0.657 & 0.609 \\

\midrule

\multirow{4}{*}{Stella}
 & RID & CSV & 0.300 & 0.543 & 0.600 & 0.543 & 0.553 \\
 & RID & Mixed & 0.243 & 0.429 & 0.500 & 0.471 & 0.515 \\
 & RID & HTML & 0.200 & 0.329 & 0.343 & 0.329 & 0.609 \\
 & RID & Markdown & 0.243 & 0.429 & 0.500 & 0.471 & 0.515 \\

\midrule

\multirow{4}{*}{Qwen3-E4B}
 & RID+Shuf & CSV & 0.271 & 0.629 & 0.686 & 0.629 & 0.432 \\
 & RID+Shuf & Mixed & 0.314 & 0.686 & 0.729 & 0.643 & 0.489 \\
 & RID+Shuf & HTML & 0.371 & 0.657 & 0.743 & 0.671 & 0.553 \\
 & RID+Shuf & Markdown & 0.343 & 0.686 & 0.743 & 0.714 & 0.480 \\

\midrule

\multirow{4}{*}{Stella}
 & RID+Shuf & CSV & 0.200 & 0.471 & 0.529 & 0.500 & 0.400 \\
 & RID+Shuf & Mixed & 0.200 & 0.400 & 0.457 & 0.429 & 0.467 \\
 & RID+Shuf & HTML & 0.100 & 0.286 & 0.329 & 0.329 & 0.304 \\
 & RID+Shuf & Markdown & 0.186 & 0.414 & 0.514 & 0.429 & 0.433 \\

\bottomrule
\end{tabular}
\end{table*}

Across formats, Ori and RID both outperform RID+Shuf in most settings (e.g., Qwen3-E4B Mixed: $0.443$/$0.371$ vs.\ $0.314$; Stella Mixed: $0.229$/$0.243$ vs.\ $0.200$). Even when row identity is made explicit, disrupting within-column value placements still degrades discrimination. This suggests that the evaluated embeddings capture shared lexical and semantic content relatively well, but remain comparatively insensitive to fine-grained row--column binding.

\subsection{Analysis of the Embedding Space}
\label{app:embedding_geometry}

We analyze the geometry of table-level embeddings as a diagnostic of the dense retrieval setting underlying the observed discrepancy between $GR@1$ and $R@1$. The goal is to characterize three aspects of the representation space: whether the global embedding inventory exhibits an obvious collapse; whether local Target$+$Sibling groups form compact semantic neighborhoods; and whether query-to-table scoring distinguishes the answerable Target from nearby Siblings. The quantities below are standard spectral summaries of embedding sets (also used in related analyses such as \citep{geometry_of_consolidation}); we use them only as diagnostics for the retrieval behavior in \ours{}, not as a geometric theory of the Semantic-Answerability Gap and not as a test of that work's consolidation theory.

\paragraph{Setup.}
We examine two scales.
\begin{itemize}
\item \textbf{Global (all 637 tables):} whether tables in the corpus are concentrated on a few shared directions or remain spread out.
\item \textbf{Local (per query):} for each query, the answerable Target together with its Sibling Distractors; whether these tables form a compact group, and whether the query score prefers the Target inside that group.
\end{itemize}
The local group is the sibling set used by $GR@k$ in the main paper. We report:
\begin{itemize}
\item $d_{\mathrm{eff}}$ (\emph{effective dimensionality}): how many independent directions carry most of the variance; smaller values indicate a thinner subspace.
\item $\bar{d}$ (\emph{mean pairwise cosine distance}): average $1-\cos$ between pairs; larger values indicate greater separation.
\item $\rho=\lambda_1/\sum_i\lambda_i$ (\emph{spectral concentration}): fraction of variance on the leading principal direction.
\item $\Delta_q=\cos(q,t_q)-\max_{s\in S_q}\cos(q,s)$: the Target--Sibling cosine margin. Positive values mean the Target wins under cosine ranking; negative values mean a Sibling scores higher.
\end{itemize}

\paragraph{Global sanity check.}
Table~\ref{tab:embedding_geometry} reports $d_{\mathrm{eff}}$ and $\bar{d}$ over all 637 tables. Mean pairwise distances are substantial ($\bar{d}\in[0.40,0.55]$), and effective dimensionality remains in the tens ($43$--$59$). The corpus-level embedding space therefore does not look like a single collapsed blob. We treat these numbers only as a sanity check: they describe the full table inventory, not the small sibling groups associated with individual queries.

\begin{table*}[htbp]
\centering
\caption{Corpus-level geometry over all 637 tables (sanity check).}
\label{tab:embedding_geometry}
\begin{tabular}{lccr}
\toprule
Model & Format & $d_{\mathrm{eff}}$ & $\bar{d}$ \\
\midrule
Qwen3-Embedding-4B & Mixed & 55.947 & 0.5257 \\
Qwen3-Embedding-4B & CSV & 45.154 & 0.4818 \\
Qwen3-Embedding-4B & HTML & 58.698 & 0.5357 \\
Qwen3-Embedding-4B & Markdown & 56.038 & 0.5394 \\
Qwen3-Embedding-0.6B & Mixed & 50.795 & 0.4943 \\
stella\_en\_1.5B\_v5 & Mixed & 43.447 & 0.5491 \\
jina-embeddings-v4 & Mixed & 49.654 & 0.3958 \\
\bottomrule
\end{tabular}
\end{table*}

\begin{table}[htbp]
\centering
\small
\caption{Local embedding geometry on Target$+$Sibling groups. $d_{\mathrm{eff}}^{\mathrm{loc}}$ and $\rho$ measure local dimensionality and spectral concentration, $\bar{d}^{\mathrm{loc}}$ is mean pairwise cosine distance, and $\Delta_q$ measures the Target's margin over its nearest Sibling.}
\label{tab:local_embedding_geometry_full}
\begin{tabular}{llccccccc}
\toprule
Model & Format & $d_{\mathrm{eff}}^{\mathrm{loc}}$ & $\rho$ & $\bar{d}^{\mathrm{loc}}$ & $\Delta_q$ & $R@1$ & $GR@1$ & $DS@1$ \\
 &  & med. & med. & med. & mean &  &  &  \\
\midrule
Qwen3-E4B & Mixed & 2.69 & 0.50 & 0.106 & $-0.024$ & 0.172 & 0.684 & 0.252 \\
Qwen3-E0.6B & Mixed & 2.79 & 0.47 & 0.113 & $-0.027$ & 0.182 & 0.627 & 0.290 \\
Qwen3-E0.6B & CSV & 2.73 & 0.49 & 0.110 & $-0.026$ & 0.139 & 0.608 & 0.228 \\
Qwen3-E0.6B & HTML & 2.70 & 0.52 & 0.084 & $-0.022$ & 0.187 & 0.651 & 0.287 \\
Qwen3-E0.6B & Markdown & 2.72 & 0.51 & 0.092 & $-0.021$ & 0.153 & 0.603 & 0.254 \\
Stella & Mixed & 2.39 & 0.57 & 0.104 & $-0.055$ & 0.096 & 0.526 & 0.182 \\
Jina & Mixed & 2.94 & 0.46 & 0.088 & $-0.021$ & 0.163 & 0.641 & 0.254 \\
\bottomrule
\end{tabular}
\end{table}

\begin{figure}[t]
\centering
\includegraphics[width=0.95\textwidth]{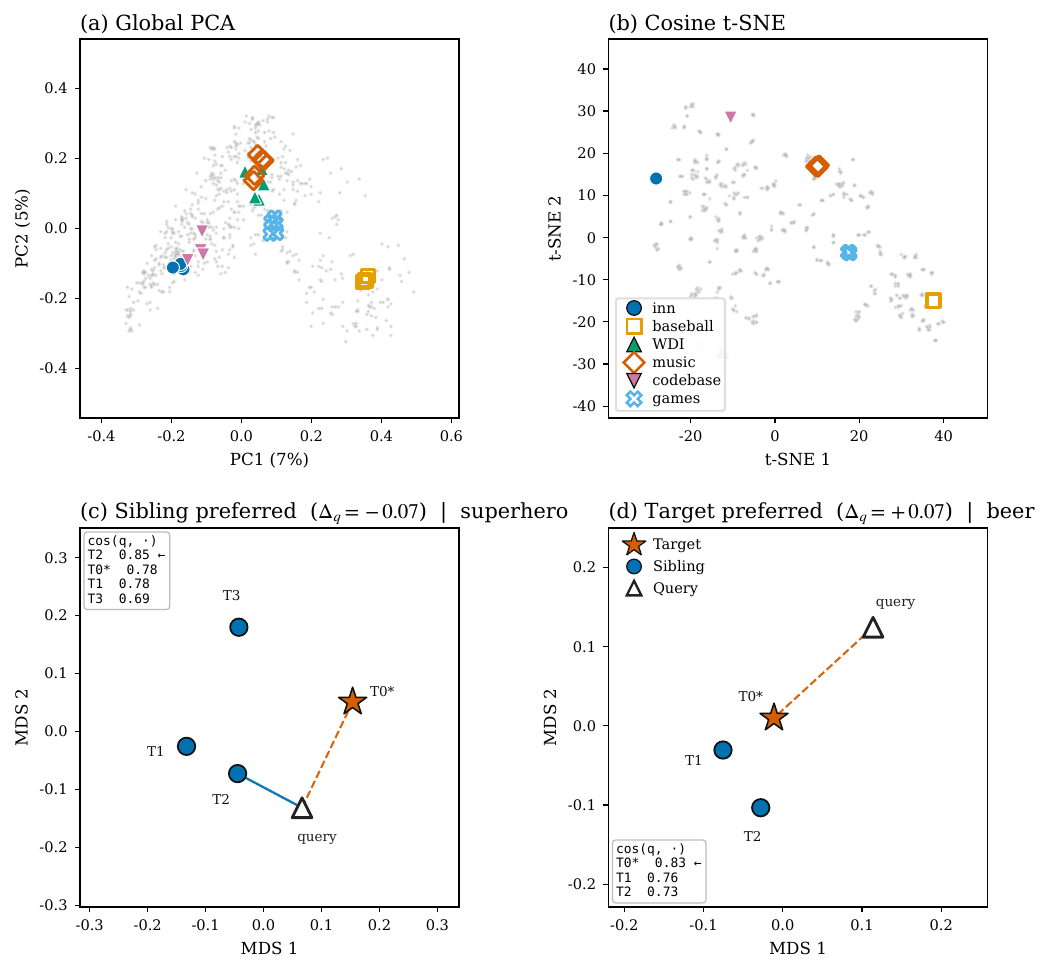}
\caption{Embedding geometry of Qwen3-Embedding-4B (Mixed) on the 637-table corpus. Colored points are six Target$+$Sibling groups; gray points are the remaining tables.
(a)~The leading two principal components explain only $12\%$ of the variance. Highlighted groups sit in different parts of a spread cloud rather than in one collapsed blob, consistent with $d_{\mathrm{eff}}\approx 56$ in Table~\ref{tab:embedding_geometry}.
(b)~A cosine t-SNE (perplexity 30) of the same points. Each highlighted group forms a tight pocket. t-SNE coordinates are a visualization of local neighborhoods, not a metric embedding.
(c, d)~Cosine-distance MDS of one sibling group together with its query. Short tags mark each table; the inset lists $\cos(q,\cdot)$ in descending order (arrow on the top-1). The dashed segment joins the query to the Target; the solid segment, when drawn, joins it to the Sibling with the higher cosine.
In (c) a Sibling outscores the Target ($\Delta_q=-0.07$); in (d) the Target wins ($\Delta_q=+0.07$).}
\label{fig:embedding_geometry}
\end{figure}

\paragraph{Local geometry and the $GR@1$ vs.\ $R@1$ gap.}
Table~\ref{tab:local_embedding_geometry_full} summarizes the per-query local analysis over $209$ queries. Sibling groups are compact: median local cosine distance is only about $0.08$--$0.11$, and median local $d_{\mathrm{eff}}$ is about $2.4$--$2.9$. Related tables therefore occupy a compact local region even though the full corpus remains spread out globally. This local structure is consistent with the retrieval behavior observed in the main results: models frequently retrieve a table from the correct sibling group without ranking the Target first. Concretely, for Mixed embeddings, $43\%$--$51\%$ of queries have $GR@1=1$ and $R@1=0$ (top-1 is a Sibling, not the Target). Figure~\ref{fig:embedding_geometry} shows the same contrast on Qwen3-Embedding-4B Mixed: the corpus stays spread in the leading principal plane, a cosine t-SNE pulls each sibling group into a tight pocket, and the sign of $\Delta_q$ appears as a within-pocket ranking between the Target and a Sibling. A compact main-text view of panels (b)--(d) appears as Figure~\ref{fig:embedding_geometry_main}.

\paragraph{Where the failure sits.}
Neighborhood shape alone does not strongly predict Target ranking: local $d_{\mathrm{eff}}$ and local $\bar{d}$ correlate only weakly with $R@1$ ($|r|\lesssim0.25$). The query-conditioned margin $\Delta_q$ is more closely aligned with this within-neighborhood scoring step: it correlates with $R@1$ at about $0.32$--$0.54$, but barely with $GR@1$. On average $\Delta_q$ is negative for every model we evaluate, and only $21\%$--$29\%$ of queries score the Target above the nearest Sibling. When the retriever reaches the neighborhood and also picks the Target, mean $\Delta_q$ is positive (e.g., $+0.025$ for Qwen3-Embedding-4B Mixed); when it reaches the neighborhood but picks a Sibling, mean $\Delta_q$ is negative ($-0.041$ on the same setting). We therefore treat $\Delta_q$ as a direct diagnostic of query-to-table scoring inside the sibling group, rather than as evidence for a new causal mechanism.

\paragraph{Scope of the local diagnosis.}
The local quantities above are defined on Target$+$Sibling Distractor groups. This controlled structure makes it possible to separate two retrieval stages that coincide under a singleton gold set: reaching a semantic neighborhood versus selecting the answerable Target within that neighborhood. In standard singleton-gold table-retrieval settings such as TARGET's FeTaQA and TabFact splits, every query has $|\mathrm{group}|=1$. Then $GR@1$ coincides with $R@1$ by construction, $DS@1=1$, the rate of ``neighborhood hit but Target miss'' is identically zero, and Target--Sibling margins as well as local $d_{\mathrm{eff}}$ on sibling groups are undefined. Table~\ref{tab:geometry_bench_contrast} reports the same dense encoder (Qwen3-Embedding-4B) under Mixed-style encoding for illustration: on TARGET FeTaQA/TabFact the metrics collapse to exact table hit, whereas on \ours{} the same model yields $GR@1=0.684$ versus $R@1=0.172$, with $51.2\%$ of queries satisfying $GR@1=1$ and $R@1=0$. Accordingly, the local quantities reported here should be understood as diagnostics enabled by the controlled retrieval structure, rather than as generic properties of table-retrieval benchmarks.

\begin{table}[htbp]
\centering
\small
\caption{Same dense encoder (Qwen3-Embedding-4B). Under singleton gold sets, group recall coincides with exact recall, so the local Target/Sibling diagnostics in this section are not defined.}
\label{tab:geometry_bench_contrast}
\begin{tabular}{lccccc}
\toprule
Benchmark & $|$group$|$ & $R@1$ & $GR@1$ & $DS@1$ & $GR@1{=}1,R@1{=}0$ \\
\midrule
TARGET FeTaQA (valid) & $1$ & $0.505$ & $0.505$ & $1.000$ & $0\%$ \\
TARGET TabFact (valid) & $1$ & $0.493$ & $0.493$ & $1.000$ & $0\%$ \\
\ours{} & $3$--$8$ & $0.172$ & $0.684$ & $0.252$ & $51.2\%$ \\
\bottomrule
\end{tabular}
\end{table}

\paragraph{Takeaway.}
Dense encoders do not simply collapse table representations into an undifferentiated space. Related tables instead form coherent local neighborhoods, as reflected by the relatively small local cosine distances together with substantial $GR@1$. The remaining difficulty is concentrated within those neighborhoods: distinguishing the answerable Target from Sibling Distractors. The Target--Sibling margin $\Delta_q$ is more directly related to this within-neighborhood scoring step than aggregate geometric statistics, providing a geometric characterization of the retrieval behavior underlying the observed $GR@1$--$R@1$ discrepancy.

\begin{figure}[t]
\centering
\includegraphics[width=0.95\textwidth]{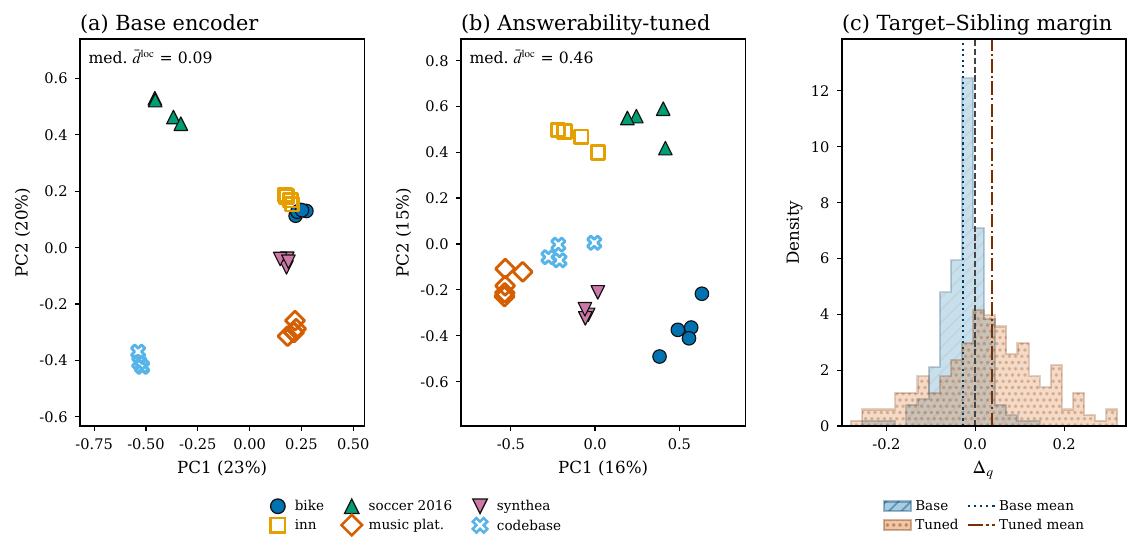}
\caption{Answerability-oriented fine-tuning changes local geometry (Qwen3-Embedding-0.6B, Mixed).
(a, b)~PCA of the same six Target$+$Sibling groups before and after tuning. Each panel is fit separately. On these groups the median within-group cosine distance rises from $0.09$ to $0.46$; across all $209$ queries the corresponding medians are $0.113$ and $0.481$.
(c)~Distribution of $\Delta_q$ on all $209$ queries. The dashed line is zero; dotted and dash-dotted lines mark the Base and Tuned means ($-0.027$ and $+0.037$).}
\label{fig:embedding_geometry_ft}
\end{figure}

Answerability-oriented fine-tuning on TCR-Bench (Appendix~\ref{app:answerability_training}) changes this picture in a way that is consistent with the same diagnostic. After tuning, mean $\Delta_q$ becomes positive and the neighborhood-hit/Target-miss rate drops, so the encoder does learn to score the Target above nearby Siblings on the training groups. At the same time local $\bar{d}$ rises sharply (median $0.113$ to $0.481$ for Qwen3-Embedding-0.6B Mixed): the compact sibling pockets of the base encoder are no longer preserved. The geometry therefore supports both sides of the training appendix---answerability is learnable inside matched sibling groups, but the update is not a drop-in repair of generic semantic retrieval. Figure~\ref{fig:embedding_geometry_ft} shows the same change for six sibling groups of Qwen3-Embedding-0.6B: the pockets that are compact before tuning spread apart afterward, and the mass of $\Delta_q$ moves from below zero to above zero.

\section{Sensitivity Analysis: Schema vs.\ Cell-Level Signals}
\label{appendix:perturbation}

This experiment provides an exploratory analysis of how the evaluated embeddings respond to schema-level and cell-level changes. We construct controlled minimal contrasts that isolate individual factors while keeping the remaining table content unchanged. The goal is to examine whether retrieval similarity responds differently to structural signals and factual cell content, providing a complementary perspective on the schema-level cue dependence observed on \ours{}.

\subsection{Experimental Design}

From the Diagnostic Subset, we generate one variant per table using three controlled perturbations: (1) \textbf{Content Perturbation (Cell-level)}, where answer-relevant cells are replaced by other valid entries from the same column while preserving the schema and table structure; (2) \textbf{Schema Perturbation (Header Rename)}, which renames a query-relevant column header to a generic token (e.g., ``Price'' $\rightarrow$ ``Column\_2'') without modifying cell values; and (3) \textbf{Schema Perturbation (Column Deletion)}, which removes the column referenced by the query to examine the effect of removing a relevant structural dimension.

These perturbations are deliberately small and isolate individual factors as cleanly as possible. This design differs from the construction of \ours{}, which uses a broader set of transformations and substantially larger contrasts between target and sibling tables. In particular, the perturbations in this appendix are minimal contrasts: only a small number of cells or a single schema component is modified while the remainder of the table is kept unchanged. The cell-level perturbation here directly replaces answer-relevant values with other entries from the same column, which can therefore change the factual identity of the affected entities. In contrast, the cell-level transformations used in \ours{} are designed to preserve the semantic identity of the underlying entities, for example through semantically similar substitutions, while introducing substantially larger changes in the table content. These differences make the two settings complementary: the present experiment isolates the effect of small, single-factor changes, whereas \ours{} evaluates retrieval under more substantial content and structural differences.

The difference is also reflected in the degree of overlap between targets and sibling tables. The minimal contrasts used here are intentionally much more similar than typical target-sibling pairs in \ours{}. Even when rows are shared in \ours{}, the maximum row-level overlap is approximately $70\%$, and such highly overlapping cases constitute only a small fraction of the benchmark. Moreover, \ours{} incorporates multiple transformation strategies rather than selecting a single dimension according to its effect on retrieval. The present experiment therefore serves as a controlled diagnostic analysis rather than a reproduction of the benchmark construction process.

\subsection{Results and Analysis}

Table~\ref{tab:perturbation_results} reports retrieval performance under these perturbations on the Mixed format. The results suggest an asymmetry in how the evaluated retrievers respond to schema-level and cell-level modifications. For example, for Qwen3-E4B, DS@1 is $0.578$ under content perturbation, compared with $0.810$ when a query-relevant header is renamed. Since a higher DS@1 indicates a stronger preference for the original target over its perturbed counterpart, the retriever is substantially more responsive to the header modification than to the corresponding cell-level modification in this controlled setting. A similar pattern is observed for Stella, where DS@1 increases from $0.611$ under content perturbation to $0.771$ under header renaming.

The cell-level perturbation results indicate that changing factual values alone can have a relatively limited effect on retrieval similarity when the surrounding schema remains unchanged. Although the perturbed tables no longer satisfy the original query conditions, their unchanged headers and structural organization can preserve substantial similarity to the target. This behavior is consistent with the failure mode studied in \ours{}, where semantically similar tables can remain difficult to distinguish when answerability depends on fine-grained factual content. However, because the perturbations here are intentionally much smaller than those in \ours{}, these results should not be interpreted as estimating the prevalence or severity of this failure mode in realistic table collections.

Column deletion provides an additional structural contrast. For Qwen3-E4B, deleting the query-relevant column produces a DS@1 of $0.738$, while Stella obtains $0.629$. These values exceed the corresponding sensitivity to content perturbation for both models, although the magnitude of the difference varies across models. Together with the header-renaming results, these observations suggest that the evaluated embedding models can be substantially affected by changes to structural or schema-level signals, even when factual cell values remain unchanged.

\subsection{Interpretation and Limitations}

The results provide preliminary evidence that the evaluated embedding representations may rely more strongly on schema-level signals than on fine-grained factual cell-level signals in minimal contrastive settings. This may explain why tables with highly similar schemas but different answer-bearing values can remain difficult to distinguish. However, we treat this observation as exploratory rather than as a definitive characterization of the underlying representations.

A key consideration is that schema tokens typically occur near the beginning of serialized tables. Consequently, modifying a header also changes information at a highly salient positional region, making positional sensitivity a potential confounding factor. Section~\ref{sec:position_bias} holds table content and answerability fixed and varies only the position of the relevant row; that analysis shows that front-loaded content receives a measurable retrieval advantage. The larger response to header perturbations therefore cannot be attributed solely to schema semantics.

This positional confound primarily affects the interpretation of \emph{why} the retrieval behavior occurs, rather than the empirical observation itself. Regardless of whether the underlying cause is schema semantics, positional salience, or their interaction, modifying relevant schema information can substantially alter retrieval behavior, whereas small factual changes can sometimes leave retrieval similarity largely unchanged. The present experiment therefore identifies a retrieval asymmetry without attempting to isolate its underlying causal mechanism.

Overall, this analysis serves as a controlled manipulation check that complements the broader benchmark results. It uses minimal single-factor contrasts to characterize model sensitivity, whereas \ours{} evaluates retrieval under more diverse and substantially larger content and structural transformations. The two settings consequently answer different questions: this experiment examines how retrieval responds to isolated changes, while \ours{} evaluates whether the tested retrievers can distinguish an answerable Target from semantically similar but non-answerable Sibling Distractors. The observed asymmetry is one possible explanation for that failure mode. The following section tests the positional confound; the precise contributions of schema information, positional effects, and cell-level representations remain open.

\begin{table*}[htbp]
\centering
\small
\caption{Retrieval performance on Minimal Contrastive Sets under small-scale perturbations (Mixed format).}
\label{tab:perturbation_results}
\begin{tabular}{l|l|ccc|c|c}
\toprule
\textbf{Configuration} & \textbf{Model} & \textbf{R@1} & \textbf{R@2} & \textbf{R@3} & \textbf{GR@1} & \textbf{DS@1} \\
\midrule
\multirow{2}{*}{\textbf{Content Perturb.}} 
& Qwen3-E4B  & 0.371 & 0.671 & 0.757 & 0.643 & 0.578 \\
& Stella & 0.314 & 0.486 & 0.529 & 0.514 & 0.611 \\
\midrule
\multirow{2}{*}{\textbf{Rename Header}} 
& Qwen3-E4B  & 0.486 & 0.671 & 0.814 & 0.600 & 0.810 \\
& Stella & 0.386 & 0.514 & 0.557 & 0.500 & 0.771 \\
\midrule
\multirow{2}{*}{\textbf{Delete Column}} 
& Qwen3-E4B  & 0.443 & 0.714 & 0.800 & 0.600 & 0.738 \\
& Stella & 0.314 & 0.429 & 0.486 & 0.500 & 0.629 \\
\bottomrule
\end{tabular}
\end{table*}

\section{Position Bias and the ``Header Preference''}
\label{sec:position_bias}

The header-renaming contrast in Section~\ref{appendix:perturbation} cannot separate schema semantics from positional salience, because headers occupy the beginning of a serialized table. We therefore test whether the evaluated embeddings exhibit \textbf{position bias} when table content and answerability are held fixed. Prior work on long-context language models has shown that performance can vary with the location of relevant information, a phenomenon commonly referred to as the \textbf{lost-in-the-middle} effect \citep{lost-in-middle}. Here we ask a narrower question: whether the physical placement of a relevant row affects retrieval similarity even when both tables remain fully answerable.

\subsection{Experimental Setup}

We construct pairs of tables that are semantically identical and \textbf{both answerable}. The only difference is the \textbf{position of the relevant row}. The row is placed in one of three positions:

\begin{itemize}
\item \textbf{Top}: immediately after the header
\item \textbf{Middle}: within the body of the table
\item \textbf{Bottom}: near the end of the table
\end{itemize}

We evaluate four comparison settings:

\begin{itemize}
\item Top vs Original (original row order in the dataset)
\item Top vs Middle
\item Bottom vs Middle
\item Top vs Bottom
\end{itemize}

For each comparison, the table listed first is treated as the formal Target for scoring, although \textbf{both tables are answerable}.

We reuse \textbf{DS@1} as a pairwise preference score. A value \textbf{greater than 0.5} indicates that the model tends to rank the \emph{first table} higher, while a value \textbf{below 0.5} indicates preference for the \emph{second table}. Thus, DS@1 here reflects positional preference rather than answerability discrimination.

\subsection{Results}

\begin{table}[htbp]
\centering
\small
\caption{Position bias evaluation under different row placements. DS@1 $>$ 0.5 indicates preference for the first table.}
\label{tab:position_bias}
\begin{tabular}{l|l|c|c|c}
\toprule
\textbf{Setting} & \textbf{Model} & \textbf{R@1} & \textbf{GR@1} & \textbf{DS@1} \\
\midrule
Top vs Original & Qwen3-E4B & 0.557 & 0.686 & 0.813 \\
Top vs Original & Stella & 0.314 & 0.514 & 0.611 \\
\midrule
Top vs Middle & Qwen3-E4B & 0.571 & 0.686 & 0.833 \\
Top vs Middle & Stella & 0.314 & 0.514 & 0.611 \\
\midrule
Bottom vs Middle & Qwen3-E4B & 0.371 & 0.629 & 0.591 \\
Bottom vs Middle & Stella & 0.300 & 0.514 & 0.583 \\
\midrule
Top vs Bottom & Qwen3-E4B & 0.557 & 0.671 & 0.830 \\
Top vs Bottom & Stella & 0.314 & 0.500 & 0.629 \\
\bottomrule
\end{tabular}
\end{table}

The results reveal a clear \textbf{header/top preference} for the two evaluated models. When the relevant row is moved to the top of the table, both tend to assign higher similarity scores to that table.

For \textbf{Qwen3-E4B}, the effect is particularly strong. In the \textit{Top vs Original} and \textit{Top vs Middle} comparisons, DS@1 reaches \textbf{0.813} and \textbf{0.833}, indicating a strong preference for tables where relevant evidence appears immediately after the header. The \textit{Top vs Bottom} setting further supports this observation (DS@1 = \textbf{0.830}): early placement increases retrieval preference under these controlled contrasts.

The \textit{Bottom vs Middle} comparison provides additional insight. When neither table places the relevant row near the header, the preference becomes weaker but still present (DS@1 = \textbf{0.591}). Earlier placement within the table body can therefore still influence similarity scores, though less strongly than header-adjacent placement.

For \textbf{Stella}, the pattern is more moderate but directionally consistent. Across all four settings, DS@1 remains between \textbf{0.58} and \textbf{0.63}, indicating a preference toward the first table in each comparison. This suggests a positional bias favoring earlier occurrences of relevant rows, although the magnitude of the effect is smaller than that observed in Qwen3-E4B.

\subsection{Discussion}

Our analysis indicates that the embedding models we evaluate exhibit a noticeable \textbf{position bias} when encoding tables. 
This observation is consistent with prior studies on long-context language models, which show that models often favor information appearing near the beginning or end of the input sequence while struggling to utilize information located in the middle \citep{lost-in-middle}. Rows placed closer to the header are more likely to influence the embedding representation and thus receive higher similarity scores during retrieval.

Notably, the preference for top-positioned rows persists even when the alternative placement is only moved to the middle of the table, indicating that positional cues can influence similarity scores in this setting.
Similar positional effects have been reported in transformer-based models, where attention patterns and positional encodings can introduce systematic biases toward certain regions of the input sequence \citep{Mitigate_position_bias}.

The position effect is weaker than the header-renaming contrast in Section~\ref{appendix:perturbation}, so positional salience does not by itself account for the full schema-versus-cell asymmetry. It does, however, show that front-loaded content receives an independent retrieval advantage. Combined with the previous section, the header-renaming result should therefore be read as a mixture of schema-level and positional effects, rather than as a pure test of schema semantics.

\subsection{Implication}

Because both tables in this experiment are answerable, the result does not measure answerability discrimination. It shows that, even with answerability held fixed, the evaluated embeddings remain sensitive to where relevant evidence appears. This positional preference is one reason header-level modifications can move retrieval more than small factual edits in our contrasts, and it is consistent with encoding tables more like linearized text than like an explicit record of row-column bindings. Primacy effects of this kind have also been reported in long-context evaluation \citep{lost-in-middle, Positional_biases_shift}.

\section{Retrieval under Table Transposition}
\label{sec:appendix_transposed_tables}

Prior studies in TableQA and table representation learning have shown that table orientation can substantially affect model behavior, especially for architectures relying on serialized table representations and positional encoding schemes \citep{table_T, table_T2, table_T3}.
Motivated by these observations, we investigate whether table transposition influences retrieval performance in our setting.

Specifically, we transpose each table by swapping rows and columns while preserving all original cell contents.
The experiment is conducted exclusively under the Mixed format setting.

\paragraph{Experimental Setup}

We evaluate representative dense embeddings on the transposed-table corpus using the same retrieval protocol and evaluation metrics as in Section~\ref{sec:main_results}.
To test whether the observed sensitivity is specific to single-vector retrieval, we additionally evaluate the row-level multi-vector approach from Appendix~\ref{sec:appendix_alternative_retrieval} under the same transposed-table setting. This method always embeds each row of the table \emph{as presented}: after transposition, a ``row'' is a row of the transposed table (i.e., an original column laid out as a row). We do not switch to column-wise indexing in order to recover the original entity rows.

\paragraph{Results}

Table~\ref{tab:transposed_results} reports the retrieval performance on transposed tables.

\begin{table*}[htbp]
\centering
\small
\caption{Retrieval performance on transposed tables under the Mixed format setting.}
\label{tab:transposed_results}
\begin{tabular}{l|ccccc}
\toprule
\textbf{Model} & \textbf{R@1} & \textbf{R@3} & \textbf{R@5} & \textbf{GR@1} & \textbf{DS@1} \\
\midrule
Qwen3-E0.6B & 0.072 & 0.211 & 0.311 & 0.311 & 0.231 \\
Qwen3-E4B & 0.081 & 0.273 & 0.392 & 0.450 & 0.181 \\
Stella & 0.038 & 0.163 & 0.258 & 0.301 & 0.127 \\
\midrule
Multi-vector (Qwen3-E0.6B) & 0.153 & 0.388 & 0.550 & 0.632 & 0.242 \\
Multi-vector (Qwen3-E4B) & 0.201 & 0.455 & 0.603 & 0.689 & 0.292 \\
\bottomrule
\end{tabular}
\end{table*}

\paragraph{Analysis}

All three evaluated single-vector embeddings show substantial degradation after table transposition.

A plausible account has two parts. First, the original tables are largely row-oriented: attributes of the same entity remain locally grouped after serialization. Many queries in \ours{} depend on such entity-level rows, so this layout can help the evaluated embeddings capture local row-level content. After transposition, related attributes become dispersed, weakening that local grouping. This is a stronger structural change than the format-level serialization variants considered in Section~\ref{sec:surface_variations}, which left Target--Sibling discrimination largely stable.

Second, transposition scatters schema information across the sequence and disrupts the alignment between values and their surrounding context. This is consistent with the header/top preference in Section~\ref{sec:position_bias}, and with limited sensitivity to row-column bindings when those associations are disrupted.

Although the evaluated row-level multi-vector retriever was comparatively strong in Appendix~\ref{sec:appendix_alternative_retrieval}, its performance also drops after transposition. Because the method still indexes the current table rows rather than the original entity rows (now columns), each vector after transposition aggregates values that no longer belong to the same entity. Its DS@1 falls to $0.242$--$0.292$, near the random-selection baseline. This suggests that the gains of this multi-vector approach depend, at least in part, on the row-oriented organization of the original tables.

Overall, this orientation stress test indicates that, for the methods we evaluate, retrieval is influenced not only by table semantics but also by structural layout. Multi-vector retrieval remains more robust than single-vector retrieval under transposition, but the drop suggests that orientation sensitivity is not limited to single-vector embeddings and can also affect row-level retrieval in this setting.

\section{Answerability-Aware Reranking}
\label{app:full_results_rerank}

\subsection{Reranking Methods}
\label{app:rerank_methods}

We evaluate three answerability-aware reranking (AAR) configurations as a diagnostic complement to dense retrieval. AAR-CE uses the Qwen3-Reranker series as a dedicated pointwise cross-encoder, with Qwen3-R0.6B, Qwen3-R4B, and Qwen3-R8B paired with Qwen3-E0.6B, Qwen3-E4B, and Qwen3-E8B, respectively. Each pair has the same parameter scale and is initialized from the same pretrained model; the two members would therefore be expected to have comparable potential, and they differ primarily in how they are used: single-vector embedding retrieval versus query-conditioned reranking. This pairing, including the 8B comparison in Section~\ref{sec:answerability_modeling}, tests whether a change in operating mode is sufficient to alter Target--Sibling discrimination. AAR-LLM-PW and AAR-LLM-LW both use Qwen3-30B. AAR-LLM-PW follows the pointwise paradigm of \citet{Pneuma}, formulating reranking as binary classification: within the reranked pool, candidates predicted as ``yes'' are placed before those predicted as ``no,'' while the relative order within each group is inherited from the dense retriever. AAR-LLM-LW instead adopts a listwise approach, feeding the query together with all candidate tables into the LLM and asking it to produce a full ordering. In all cases, reranking is applied to the top $k$ candidates from the dense retriever, denoted T$k$; T$\infty$ reranks the full retrieved list.

\begin{table*}[tp]
\centering
\small
\caption{Complete experimental results across retrieval configurations. Here, T$i$ indicates that reranking is applied to the top $i$ results returned by the embedding-based retrieval model.}
\label{tab:full-results_rerank}
\resizebox{\textwidth}{!}{
\begin{tabular}{l|l|ccccc|ccccc|c}
\toprule
\multirow{2}{*}{\textbf{Method}} & \multirow{2}{*}{\textbf{Retrieval Model}} 
& \multicolumn{5}{c|}{\textbf{R@k}} & \multicolumn{5}{c|}{\textbf{GR@k}} & \multirow{2}{*}{\textbf{DS@1}} \\
\cmidrule(lr){3-7} \cmidrule(lr){8-12}
& & \textbf{@1} & \textbf{@2} & \textbf{@3} & \textbf{@4} & \textbf{@5} 
& \textbf{@1} & \textbf{@2} & \textbf{@3} & \textbf{@4} & \textbf{@5} & \\
\midrule
\multirow{10}{*}{Qwen3-E0.6B}
 & AAR-CE (T3) & 0.273 & 0.383 & 0.411 & - & - & 0.694 & 0.665 & 0.593 & - & - & 0.393 \\
 & AAR-CE (T5) & 0.297 & 0.435 & 0.507 & 0.545 & 0.555 & 0.737 & 0.703 & 0.641 & 0.596 & 0.556 & 0.403 \\
 & AAR-CE (T10) & 0.359 & 0.541 & 0.617 & 0.665 & 0.689 & 0.789 & 0.761 & 0.719 & 0.695 & 0.664 & 0.455 \\
 & AAR-CE (T20) & 0.416 & 0.589 & 0.684 & 0.732 & 0.770 & 0.823 & 0.797 & 0.758 & 0.744 & 0.736 & 0.506 \\
 & AAR-CE (T50) & 0.459 & 0.632 & 0.742 & 0.799 & 0.833 & 0.876 & 0.835 & 0.797 & 0.782 & 0.772 & 0.525 \\
 & AAR-CE (T$\infty$) & 0.507 & 0.694 & 0.804 & 0.866 & 0.900 & 0.895 & 0.861 & 0.818 & 0.794 & 0.771 & 0.567 \\
 & AAR-LLM-PW (T5) & 0.383 & 0.455 & 0.512 & 0.545 & 0.555 & 0.689 & 0.639 & 0.600 & 0.574 & 0.556 & 0.556 \\
 & AAR-LLM-PW (T10) & 0.469 & 0.589 & 0.641 & 0.675 & 0.684 & 0.751 & 0.694 & 0.633 & 0.603 & 0.589 & 0.624 \\
 & AAR-LLM-LW (T5) & 0.383 & 0.445 & 0.488 & 0.522 & 0.555 & 0.737 & 0.689 & 0.630 & 0.600 & 0.584 & 0.519 \\
 & AAR-LLM-LW (T10) & 0.426 & 0.493 & 0.550 & 0.603 & 0.646 & 0.751 & 0.730 & 0.689 & 0.666 & 0.658 & 0.567 \\
\midrule
\multirow{6}{*}{Qwen3-E4B}
 & AAR-CE (T5) & 0.426 & 0.522 & 0.589 & 0.617 & 0.632 & 0.789 & 0.749 & 0.711 & 0.682 & 0.662 & 0.539 \\
 & AAR-CE (T10) & 0.498 & 0.598 & 0.651 & 0.694 & 0.722 & 0.828 & 0.801 & 0.777 & 0.757 & 0.735 & 0.601 \\
 & AAR-LLM-PW (T5) & 0.445 & 0.550 & 0.598 & 0.612 & 0.632 & 0.746 & 0.725 & 0.692 & 0.673 & 0.662 & 0.596 \\
 & AAR-LLM-PW (T10) & 0.517 & 0.622 & 0.689 & 0.703 & 0.718 & 0.804 & 0.766 & 0.721 & 0.688 & 0.676 & 0.643 \\
 & AAR-LLM-LW (T5) & 0.474 & 0.522 & 0.550 & 0.603 & 0.632 & 0.770 & 0.737 & 0.713 & 0.692 & 0.679 & 0.615 \\
 & AAR-LLM-LW (T10) & 0.435 & 0.526 & 0.593 & 0.632 & 0.684 & 0.804 & 0.792 & 0.767 & 0.743 & 0.733 & 0.542 \\
\midrule
\multirow{6}{*}{Qwen3-E8B}
 & AAR-CE (T5) & 0.507 & 0.608 & 0.636 & 0.660 & 0.665 & 0.770 & 0.730 & 0.684 & 0.653 & 0.622 & 0.658 \\
 & AAR-CE (T10) & 0.574 & 0.632 & 0.651 & 0.656 & 0.656 & 0.766 & 0.732 & 0.708 & 0.683 & 0.664 & 0.750 \\ \\
 & AAR-LLM-PW (T5) & 0.498 & 0.608 & 0.641 & 0.660 & 0.665 & 0.732 & 0.694 & 0.652 & 0.629 & 0.622 & 0.680 \\
 & AAR-LLM-PW (T10) & 0.536 & 0.651 & 0.684 & 0.722 & 0.737 & 0.761 & 0.720 & 0.681 & 0.650 & 0.637 & 0.704 \\
 & AAR-LLM-LW (T5) & 0.455 & 0.555 & 0.593 & 0.646 & 0.665 & 0.756 & 0.718 & 0.691 & 0.657 & 0.642 & 0.601 \\
 & AAR-LLM-LW (T10) & 0.450 & 0.555 & 0.593 & 0.670 & 0.708 & 0.770 & 0.742 & 0.718 & 0.700 & 0.698 & 0.584 \\
\bottomrule
\end{tabular}
}
\end{table*}

\subsection{Results and Analysis}
\label{app:rerank_analysis}

Table~\ref{tab:full-results_rerank} reports the complete results across retrieval models and reranking configurations. Across the evaluated settings, AAR improves target identification over the corresponding dense ranking. The improvement is particularly clear for R@1. For the Qwen3-E0.6B and Qwen3-R0.6B pair, Mixed-format dense R@1 is 0.182; AAR-CE raises it to 0.297 at T5, 0.359 at T10, and 0.507 when all retrieved candidates are reranked. The same-scale pattern also holds at 4B and 8B: AAR-CE (T10) reaches 0.498 and 0.574, respectively, well above the dense Mixed-format R@1 of 0.172 and 0.182. Because each embedding/reranker pair shares scale and initialization, this gap is difficult to attribute to extra capacity. It indicates that changing the operating mode---from single-vector similarity to query-conditioned answerability assessment---substantially improves Target ranking once the Target is included in the reranked pool.

For AAR-CE on this pair, larger reranking budgets continue to help. As $k$ increases from 5 to 50, R@5 rises from 0.555 to 0.833, while R@1 increases from 0.297 to 0.459. GR@1 also increases from 0.737 to 0.876, and DS@1 rises monotonically from 0.393 at T3 to 0.567 at T$\infty$. The joint increase in R@1, GR@1, and DS@1 suggests that exposing more candidates to the reranker improves both neighborhood retrieval and within-group target identification, rather than trading one for the other. Gains remain substantial beyond T10, suggesting that restricting AAR-CE to a very small pool leaves answerable candidates unused. Listwise LLM reranking does not follow the same monotonic pattern: for Qwen3-E4B and Qwen3-E8B, AAR-LLM-LW at T10 yields lower R@1 than at T5.

\begin{figure}[t]
    \centering
    \includegraphics[width=0.8\linewidth]{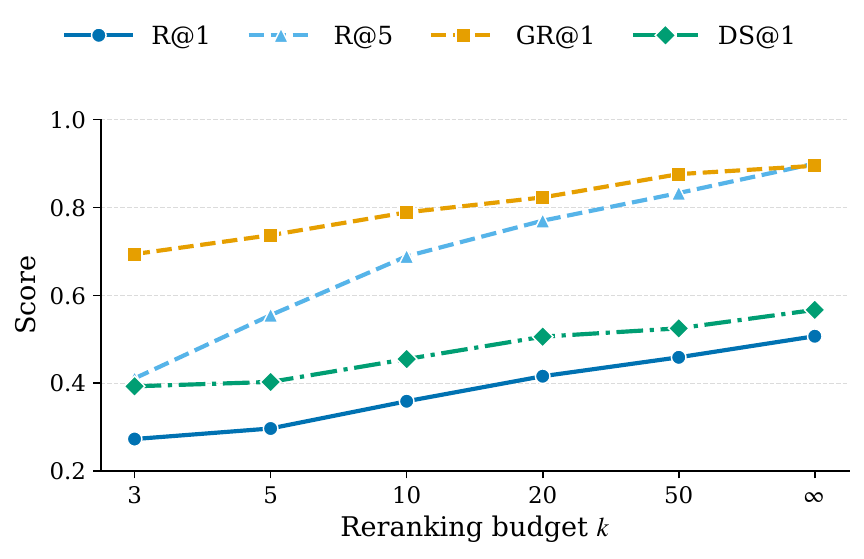}
    \caption{Scaling behavior of answerability-aware reranking with different reranking budgets $k$, using Qwen3-E0.6B for initial retrieval and Qwen3-R0.6B for reranking. For T3, R@5 is effectively R@3 because only the top-3 candidates are reranked.}
    \label{fig:rerank-k-scaling}
\end{figure}

Figure~\ref{fig:rerank-k-scaling} visualizes this scaling for the Qwen3-E0.6B and Qwen3-R0.6B pair. Retrieval improves as more candidates are reranked, but the gains diminish at larger $k$. This gives an empirical trade-off between reranking budget and retrieval quality for this configuration.

Despite these gains, all evaluated reranking configurations are substantially more expensive than embedding retrieval. AAR-CE requires an additional cross-encoder forward pass for each reranked candidate, AAR-LLM-PW requires one LLM call per candidate, and AAR-LLM-LW jointly processes all candidates in a long context. We therefore use T$k$ as the default setting for controlled comparisons and report T$\infty$ only for AAR-CE with the Qwen3-E0.6B and Qwen3-R0.6B pair. This setting reranks all retrieved candidates and serves as an empirical upper bound on the benefit obtainable from AAR-CE at this scale, rather than as an exhaustive evaluation across all model scales and reranking configurations.

\section{Answerability-Oriented Training}
\label{app:answerability_training}
The reranking experiments show that explicit query-conditioned interaction can substantially improve target identification. We next investigate whether an embedding model can learn a similar answerability signal through direct supervision. These experiments are diagnostic and exploratory rather than conventional held-out evaluations, since neither TCR-Bench nor Real-TCR provides a dedicated training split. Our goal is to examine whether answerability can be learned and whether the resulting behavior transfers across table distributions while preserving general retrieval capability. We emphasize that the training setup here serves this exploratory purpose rather than aiming to establish a standardized or optimized training procedure; the experiments are intended as a diagnostic probe, not as a benchmark-oriented training recipe.

\subsection{Experimental Setup}
\label{app:answerability_training_setup}

We fine-tune Qwen3-Embedding-0.6B using query--target positives and sibling-table hard negatives. The positive pair consists of the query and its target table, while sibling tables provide semantically related but non-target negatives. We use a maximum of 16 negatives per query.
We first conduct preliminary experiments to determine the remaining training settings (including temperature) and then perform a learning-rate grid search. Based on these preliminary runs, the temperature is fixed to 0.1 throughout the search. The learning-rate candidates are $1\times10^{-5}$, $3\times10^{-5}$, $1\times10^{-4}$, $2\times10^{-4}$, $3\times10^{-4}$, $5\times10^{-4}$, $1\times10^{-3}$, $2\times10^{-3}$, and $5\times10^{-3}$. Other hyperparameters such as maximum sequence length and batch size are chosen under computational constraints.
Unless noted otherwise, we report R@1, and on TCR-Bench / Real-TCR also GR@1 and DS@1 (DS@1 $=\mathrm{R@1}/\mathrm{GR@1}$ at the aggregate level, following the main evaluation).

\subsection{Training on TCR-Bench}
\label{app:tcr_answerability_training}

The best TCR-Bench configuration uses a learning rate of $1\times10^{-4}$. We train for up to five epochs, with the best checkpoint obtained at epoch 4.

\begin{table}[htbp]
\centering
\small
\caption{Training configuration for answerability-oriented fine-tuning on TCR-Bench.}
\label{tab:answerability_training_config}
\begin{tabular}{ll}
\toprule
\textbf{Configuration} & \textbf{Value} \\
\midrule
Base model & Qwen3-Embedding-0.6B \\
Learning rate & $1\times10^{-4}$ \\
Temperature & 0.1 \\
Epochs & 5 (best: epoch 4) \\
Batch size & 8 \\
Gradient accumulation & 2 \\
Maximum length (query / passage) & 512 / 512 \\
Maximum negatives & 16 \\
Warmup ratio & 0.05 \\
Weight decay & 0.01 \\
Mixed precision & bf16 \\
Seed & 42 \\
\bottomrule
\end{tabular}
\end{table}

The best checkpoint reaches training R@1 of 0.493 with an average epoch loss of 0.0669. On TCR-Bench, it achieves R@1 of 0.493, GR@1 of 0.713, and DS@1 of 0.691.

\begin{table}[htbp]
\centering
\small
\caption{Answerability-oriented fine-tuning on TCR-Bench.}
\label{tab:answerability_tcr_training}
\begin{tabular}{lccc}
\toprule
\textbf{Model} & \textbf{R@1} & \textbf{GR@1} & \textbf{DS@1} \\
\midrule
Qwen3-Embedding-0.6B & 0.182 & 0.627 & 0.290 \\
Answerability-tuned & \textbf{0.493} & \textbf{0.713} & \textbf{0.691} \\
\bottomrule
\end{tabular}
\end{table}

The large improvement in R@1 and DS@1 shows that direct answerability-oriented supervision can substantially alter retrieval behavior under a matched distribution. This provides evidence that answerability is learnable for the embedding model we fine-tune, rather than inaccessible under sibling hard-negative supervision.

\subsubsection{Geometry of the tuned encoder}
We repeat the Target--Sibling diagnostics of Appendix~\ref{app:embedding_geometry} on this checkpoint (Mixed encodings of the same 637 tables and 209 queries). Table~\ref{tab:answerability_geometry} compares the unmatched Qwen3-Embedding-0.6B encoder with the TCR-Bench-tuned model. The Target--Sibling margin $\Delta_q$ changes sign (mean $-0.027$ to $+0.037$), the fraction of queries with $\Delta_q>0$ rises from $27\%$ to $63\%$, and the neighborhood-hit/Target-miss rate falls from $44.5\%$ to $22.0\%$. When the tuned retriever reaches the sibling group and ranks the Target first, mean $\Delta_q$ is $+0.122$; when it reaches the group but ranks a Sibling first, mean $\Delta_q$ is $-0.094$. Query-to-table scoring inside the training sibling groups therefore does become answerability-sensitive, which is the geometric counterpart of the DS@1 gain.

The same update does not preserve the compact local neighborhoods of the base encoder. Median local pairwise cosine distance rises from $0.113$ to $0.481$, and corpus-level $\bar{d}$ rises from $0.494$ to $0.952$. Sibling tables that previously occupied a tight pocket are pushed far apart. The encoder learns an answerability-sensitive scoring rule on the groups it was trained on, but it does so by disrupting the semantic neighborhood structure that generic retrieval relies on.

\begin{table}[htbp]
\centering
\small
\caption{Target--Sibling geometry on TCR-Bench before and after answerability-oriented fine-tuning (Mixed). $\bar{d}^{\mathrm{loc}}$ is the median local pairwise cosine distance; $\Delta_q$ is the Target--Sibling cosine margin.}
\label{tab:answerability_geometry}
\begin{tabular}{lcccccc}
\toprule
\textbf{Model} & $\bar{d}^{\mathrm{loc}}$ & $\Delta_q$ & $\Delta_q{>}0$ & $R@1$ & $GR@1$ & $DS@1$ \\
 & med. & mean &  &  &  &  \\
\midrule
Qwen3-Embedding-0.6B & 0.113 & $-0.027$ & $27\%$ & 0.182 & 0.627 & 0.290 \\
Answerability-tuned & 0.481 & $+0.037$ & $63\%$ & 0.493 & 0.713 & 0.691 \\
\bottomrule
\end{tabular}
\end{table}

\subsubsection{Cross-Distribution Evaluation}
\label{app:answerability_cross_distribution}

We next evaluate whether this in-domain answerability signal transfers. We consider transfer between TCR-Bench and Real-TCR, as well as Real-TCR-Small. Since neither benchmark provides a dedicated training split, these experiments are controlled transfer diagnostics rather than standard held-out evaluations.

\begin{table}[htbp]
\centering
\small
\caption{Cross-distribution evaluation of answerability-oriented fine-tuning. The third row selects the best TCR-Bench-trained checkpoint by Real-TCR R@1 (full-parameter grid; LoRA ties at the same R@1). GR@1 and DS@1 are omitted for that row because the selection pass reported only R@1.}
\label{tab:answerability_cross_distribution}
\begin{tabular}{llcccc}
\toprule
\textbf{Training} & \textbf{Evaluation} & \textbf{Selection} & \textbf{R@1} & \textbf{GR@1} & \textbf{DS@1} \\
\midrule
TCR-Bench & Real-TCR & TCR-Bench R@1 & 0.065 & 0.831 & 0.078 \\
TCR-Bench & Real-TCR-Small & TCR-Bench R@1 & 0.132 & 0.868 & 0.152 \\
TCR-Bench & Real-TCR & Real-TCR R@1 & 0.156 & --- & --- \\
\midrule
Real-TCR & Real-TCR & Real-TCR R@1 & \textbf{0.442} & \textbf{0.922} & \textbf{0.479} \\
Real-TCR & TCR-Bench & Real-TCR R@1 & 0.043 & 0.311 & 0.138 \\
\bottomrule
\end{tabular}
\end{table}

The TCR-Bench in-domain checkpoint transfers poorly: 0.065 R@1 and 0.078 DS@1 on Real-TCR, and 0.132 R@1 and 0.152 DS@1 on Real-TCR-Small. Because that evaluation uses a checkpoint selected on TCR-Bench, we additionally sweep the same training run (full-parameter and LoRA learning-rate grids) and retain the checkpoint with the highest Real-TCR R@1. The best transferred configuration reaches only 0.156 R@1 on Real-TCR---above the unmatched 0.6B baseline of 0.078, but far from matched Real-TCR training (0.442). LoRA matches this same Real-TCR R@1 while remaining close to the base encoder on TCR-Bench (0.191 R@1, 0.270 DS@1) and on TARGET, so the bump does not require a stable transferred answerability rule. Target-set selection therefore yields only a modest, brittle gain rather than a ready-made transferable recipe under this setup. In the reverse direction, Real-TCR training produces strong within-distribution performance (0.442 R@1, 0.479 DS@1) but drops to 0.043 R@1 and 0.138 DS@1 on TCR-Bench. The learned answerability behavior remains substantially distribution-dependent.

\subsubsection{Effect on General Retrieval}
\label{app:answerability_general_retrieval}
We further evaluate the TCR-Bench-trained model on independently sampled subsets of TabFact and FeTaQA. These benchmarks do not define candidate groups, so only R@1 is reported.

\begin{table}[htbp]
\centering
\small
\caption{Retrieval performance of the TCR-Bench-trained model on independent TARGET subsets.}
\label{tab:target_ood}
\begin{tabular}{lc}
\toprule
\textbf{Benchmark} & \textbf{R@1} \\
\midrule
\textsc{TabFact}-valid sample & 0.184 \\
\textsc{TabFact}-test sample & 0.179 \\
\textsc{FeTaQA}-valid sample & 0.152 \\
\textsc{FeTaQA}-test sample & 0.104 \\
\bottomrule
\end{tabular}
\end{table}

The TCR-Bench-trained model achieves R@1 of 0.184 and 0.179 on the TabFact validation and test samples, respectively, and 0.152 and 0.104 on the corresponding FeTaQA samples. Compared with the base-model results of 0.414 on TabFact-valid and 0.325 on FeTaQA-valid, these results indicate a substantial loss in broad semantic retrieval capability after answerability-oriented fine-tuning. The Real-TCR-selected checkpoint from the same run is no better on these TARGET subsets (0.147 / 0.152 on TabFact valid/test and 0.133 / 0.096 on FeTaQA valid/test). This is consistent with the geometry above: spreading sibling neighborhoods on TCR-Bench does not yield a retrieval function that remains useful on unrelated table collections.

\subsection{Training on General Retrieval Data}
\label{app:target_mix_training}

The previous results show that answerability-oriented fine-tuning on TCR-Bench substantially degrades retrieval on TabFact and FeTaQA. We now consider the reverse direction: whether strengthening generic table retrieval on these TARGET collections improves answerability-sensitive ranking on TCR-Bench and Real-TCR.

We construct a 1:1 mixture from the TARGET TabFact and FeTaQA training splits, using 7{,}326 examples from each source (14{,}652 in total), with one TabFact query sampled per table. Table serializations follow the evaluation default (HTML for TabFact, JSON for FeTaQA), and each example uses the corresponding evaluation query prefix (fact verification vs.\ TableQA). Validation queries are excluded from training. Unlike TCR-Bench and Real-TCR, these TARGET collections do not define sibling groups, so the objective is query--table relevance with in-batch negatives rather than target-versus-sibling discrimination.

We fine-tune Qwen3-Embedding-0.6B with LoRA ($r=16$, $\alpha=32$) using the same contrastive recipe and temperature 0.1. The learning-rate grid is $5\times10^{-5}$, $1\times10^{-4}$, $2\times10^{-4}$, $3\times10^{-4}$, and $5\times10^{-4}$. Checkpoints are selected independently on the TabFact-valid and FeTaQA-valid samples used in the main evaluation.

\begin{table}[htbp]
\centering
\small
\caption{Training configuration for general-retrieval fine-tuning on the TARGET TabFact+FeTaQA mix. The two learning rates correspond to the TabFact-valid and FeTaQA-valid selections, respectively.}
\label{tab:target_mix_training_config}
\begin{tabular}{ll}
\toprule
\textbf{Configuration} & \textbf{Value} \\
\midrule
Base model & Qwen3-Embedding-0.6B \\
Training data & TARGET TabFact+FeTaQA 1:1 mix (14{,}652) \\
Adaptation & LoRA ($r=16$, $\alpha=32$, dropout 0.05) \\
Learning rate & $3\times10^{-4}$ / $2\times10^{-4}$ (see text) \\
Temperature & 0.1 \\
Epochs & 5 (best: epoch 3 for both selections) \\
Batch size & 4 \\
Gradient accumulation & 4 \\
Maximum length (query / passage) & 512 / 2048 \\
Maximum negatives & 16 \\
Warmup ratio & 0.05 \\
Weight decay & 0.01 \\
Mixed precision & bf16 \\
Seed & 42 \\
\bottomrule
\end{tabular}
\end{table}

The TabFact-selected checkpoint uses a learning rate of $3\times10^{-4}$ and reaches R@1 of 0.459 on TabFact-valid. The FeTaQA-selected checkpoint uses $2\times10^{-4}$ and reaches R@1 of 0.408 on FeTaQA-valid. Both improve over the corresponding base-model results of 0.414 and 0.325. As above, TabFact and FeTaQA do not define sibling groups, so GR@1 and DS@1 are undefined and only R@1 is reported.

\begin{table}[htbp]
\centering
\small
\caption{In-domain R@1 of TARGET-mix LoRA checkpoints on the TabFact and FeTaQA validation samples.}
\label{tab:target_mix_indomain}
\begin{tabular}{lcc}
\toprule
\textbf{Model} & \textbf{TabFact-valid} & \textbf{FeTaQA-valid} \\
\midrule
Qwen3-Embedding-0.6B & 0.414 & 0.325 \\
TARGET mix (TabFact-best) & \textbf{0.459} & 0.370 \\
TARGET mix (FeTaQA-best) & 0.453 & \textbf{0.408} \\
\bottomrule
\end{tabular}
\end{table}

We then evaluate both checkpoints on TCR-Bench and Real-TCR, where sibling groups are defined and R@1, GR@1, and DS@1 are all reported.

\begin{table}[htbp]
\centering
\small
\caption{Transfer of TARGET-mix general-retrieval training to answerability benchmarks, compared with the unmatched base encoder and matched answerability-oriented fine-tuning.}
\label{tab:target_mix_transfer}
\begin{tabular}{llcccc}
\toprule
\textbf{Training} & \textbf{Evaluation} & \textbf{Model} & \textbf{R@1} & \textbf{GR@1} & \textbf{DS@1} \\
\midrule
--- & TCR-Bench & Qwen3-Embedding-0.6B & 0.182 & 0.627 & 0.290 \\
TARGET mix & TCR-Bench & TabFact-best & 0.100 & 0.450 & 0.223 \\
TARGET mix & TCR-Bench & FeTaQA-best & 0.163 & 0.555 & 0.293 \\
TCR-Bench & TCR-Bench & Answerability-tuned & \textbf{0.493} & \textbf{0.713} & \textbf{0.691} \\
\midrule
--- & Real-TCR & Qwen3-Embedding-0.6B & 0.078 & 0.870 & 0.090 \\
TARGET mix & Real-TCR & TabFact-best & 0.039 & 0.857 & 0.045 \\
TARGET mix & Real-TCR & FeTaQA-best & 0.091 & 0.896 & 0.101 \\
Real-TCR & Real-TCR & Answerability-tuned & \textbf{0.442} & \textbf{0.922} & \textbf{0.479} \\
\bottomrule
\end{tabular}
\end{table}

Neither TARGET-mix checkpoint recovers the answerability-oriented in-domain gains. On TCR-Bench, the FeTaQA-selected model remains closest to the base encoder: R@1 of 0.163 versus 0.182, GR@1 of 0.555 versus 0.627, and DS@1 of 0.293 versus 0.290. Neighborhood retrieval declines, while within-neighborhood discrimination is essentially unchanged. The TabFact-selected model degrades on all three metrics (0.100 / 0.450 / 0.223). On Real-TCR, both mix checkpoints retain high GR@1 (0.857 and 0.896 versus the base 0.870), but DS@1 stays near the base (0.045 and 0.101 versus 0.090). Matched answerability-oriented training instead reaches 0.493 / 0.713 / 0.691 on TCR-Bench and 0.442 / 0.922 / 0.479 on Real-TCR, with the largest relative gain in DS@1. Under this contrastive setup, generic query--table relevance training does not by itself induce the target-versus-sibling discrimination that answerability benchmarks require: models can still enter the correct semantic neighborhood while remaining unable to identify the answerable table. Together with the drop on TabFact and FeTaQA after TCR-Bench training, the two objectives are not interchangeable in a simple way.

\subsection{Joint Training on TCR-Bench and TARGET}
\label{app:joint_tcr_target_mix}

The preceding experiments examine the two directions separately. We next ask whether a single mixed training run can support both answerability-oriented and generic retrieval signals at once. This is again a diagnostic probe rather than an attempt to optimize a production recipe: if a straightforward mixture already aligned the two objectives, the Semantic-Answerability Gap would be easier to close through data mixing alone.

We concatenate the TARGET TabFact+FeTaQA mix above with TCR-Bench hard-negative examples. Because TCR-Bench contributes only 209 queries, concatenating it as-is would leave answerability supervision at about $1.4\%$ of the mix; we therefore upsample TCR-Bench (with replacement, keeping sibling hard negatives) to 7{,}326 examples so that each epoch is approximately 1:1:1 TabFact / FeTaQA / TCR-Bench (21{,}978 examples total). Query prefixes and table serializations follow the same eval-aligned settings as Section~\ref{app:target_mix_training}. We fine-tune with LoRA ($r=16$, $\alpha=32$) and a reduced learning-rate grid of $1\times10^{-4}$ and $2\times10^{-4}$ (rates that previously collapsed TARGET under full-parameter updates, or moved little under LoRA, are omitted). After each epoch we record R@1 on Real-TCR (sibling metrics GR@1/DS@1 are not used for selection here) and on the TabFact-valid and FeTaQA-valid samples (TARGET has no sibling groups, so only R@1 applies), and keep a separate best checkpoint per bench.

\begin{table}[htbp]
\centering
\small
\caption{Training configuration for joint TCR-Bench + TARGET-mix LoRA fine-tuning.}
\label{tab:joint_mix_training_config}
\begin{tabular}{ll}
\toprule
\textbf{Configuration} & \textbf{Value} \\
\midrule
Base model & Qwen3-Embedding-0.6B \\
Training data & TCR-Bench + TabFact+FeTaQA (1:1:1, 21{,}978) \\
Adaptation & LoRA ($r=16$, $\alpha=32$, dropout 0.05) \\
Learning rate & $1\times10^{-4}$ / $2\times10^{-4}$ (see text) \\
Temperature & 0.1 \\
Epochs & 5 \\
Batch size & 4 \\
Gradient accumulation & 4 \\
Maximum length (query / passage) & 512 / 2048 \\
Maximum negatives & 16 \\
Query format & eval\_prefix \\
\bottomrule
\end{tabular}
\end{table}

\begin{table}[htbp]
\centering
\small
\caption{R@1 under joint TCR-Bench+TARGET LoRA training, compared with the base encoder, TARGET-mix-only LoRA, and matched Real-TCR training. Joint-mix rows use the checkpoint best on the named selection bench; TARGET-mix-only Real-TCR uses the FeTaQA-best transfer from Table~\ref{tab:target_mix_transfer}.}
\label{tab:joint_mix_results}
\begin{tabular}{lccc}
\toprule
\textbf{Model} & \textbf{Real-TCR} & \textbf{TabFact-valid} & \textbf{FeTaQA-valid} \\
\midrule
Qwen3-Embedding-0.6B & 0.078 & 0.414 & 0.325 \\
TARGET mix only (best per bench) & 0.091 & \textbf{0.459} & \textbf{0.408} \\
Joint mix (Real-TCR-best) & \textbf{0.117} & 0.374 & 0.367 \\
Joint mix (TabFact-best) & 0.065 & 0.428 & 0.382 \\
Joint mix (FeTaQA-best) & 0.078 & 0.422 & 0.393 \\
Real-TCR matched training & 0.442 & --- & --- \\
\bottomrule
\end{tabular}
\end{table}

The mixture does not remove the tradeoff. Selecting on Real-TCR yields only 0.117 R@1---above the 0.078 baseline and the TARGET-mix-only transfer of 0.091, but far from matched Real-TCR training (0.442), and at the cost of TabFact-valid dropping to 0.374 (below the 0.414 base). Selecting on TabFact or FeTaQA recovers moderate TARGET gains (0.428 / 0.393), yet these remain below the TARGET-mix-only peaks (0.459 / 0.408), and Real-TCR stays near the base (0.065 / 0.078). No single checkpoint in the completed LoRA grid jointly exceeds the separate single-objective peaks on both sides. Under this exploratory mixture, therefore, simply pooling answerability hard negatives with generic TARGET relevance data does not yield an easy joint optimum.

\subsection{Training on Real-TCR}
\label{app:real_tcr_answerability_training}
We additionally fine-tune Qwen3-Embedding-0.6B directly on Real-TCR using the available hard-negative training data. The best configuration uses a learning rate of $1\times10^{-4}$ and temperature 0.1, with the best checkpoint obtained at epoch 2 of 5.
\begin{table}[htbp]
\centering
\small
\caption{Training configuration for Real-TCR answerability-oriented fine-tuning.}
\label{tab:real_tcr_training_config}
\begin{tabular}{ll}
\toprule
\textbf{Configuration} & \textbf{Value} \\
\midrule
Base model & Qwen3-Embedding-0.6B \\
Learning rate & $1\times10^{-4}$ \\
Temperature & 0.1 \\
Epochs & 5 (best: epoch 2) \\
Batch size & 8 \\
Gradient accumulation & 2 \\
Query / passage maximum length & 512 / 512 \\
Maximum negatives & 16 \\
Warmup ratio & 0.05 \\
Weight decay & 0.01 \\
Mixed precision & bf16 \\
Seed & 42 \\
\bottomrule
\end{tabular}
\end{table}

The best checkpoint achieves training R@1 of 0.442 on the full Real-TCR training collection, with an average epoch loss of 0.905.

\begin{table}[htbp]
\centering
\small
\caption{Evaluation of the Real-TCR-trained model across table-retrieval distributions.}
\label{tab:real_tcr_training_results}
\begin{tabular}{lccc}
\toprule
\textbf{Benchmark} & \textbf{R@1} & \textbf{GR@1} & \textbf{DS@1} \\
\midrule
Real-TCR & 0.442 & 0.922 & 0.479 \\
Real-TCR-Small & 0.711 & 0.947 & 0.750 \\
TCR-Bench & 0.043 & 0.311 & 0.138 \\
\bottomrule
\end{tabular}
\end{table}

Matched Real-TCR training remains the only setting in which Real-TCR R@1 rises substantially (0.078 to 0.442); the joint TCR-Bench+TARGET mixture above reaches at most 0.117. The matched gain does not transfer to TCR-Bench (0.043 R@1, 0.138 DS@1), consistent with the reverse of the modest Real-TCR-selected transfer above.

\subsection{Discussion}
\label{app:answerability_training_discussion}

Taken together, these experiments provide a nuanced interpretation of the Semantic-Answerability Gap under the diagnostic setups above. Direct supervision substantially improves answerability-sensitive retrieval on the training distribution, and the geometry of the tuned encoder confirms that this is not a bookkeeping artifact: $\Delta_q$ becomes positive on a majority of TCR-Bench queries, and DS@1 rises with it. Answerability is therefore learnable under matched sibling supervision.

That learned behavior, however, does not by itself resolve the gap. Transfer across table distributions remains weak even when we select the TCR-Bench-trained checkpoint by Real-TCR R@1 (0.156 versus a 0.078 baseline and 0.442 matched training). The same update spreads local sibling neighborhoods and degrades generic retrieval on TabFact and FeTaQA. The reverse is also true: strengthening generic table retrieval on a TARGET TabFact+FeTaQA mixture improves in-domain R@1 on those collections, yet leaves DS@1 near the base encoder on TCR-Bench and Real-TCR, even when GR@1 remains high. Mixing the two data sources in one LoRA run likewise fails to deliver an easy joint win: Real-TCR R@1 rises only modestly (to 0.117), while TARGET peaks fall short of TARGET-only training, and the best checkpoint on one side is not best on the other.

These results do not claim that no joint objective can succeed. They show that, under the straightforward contrastive recipes explored here, fitting an answerability signal, transferring it, and preserving broad semantic relevance do not come for free together. A hard-negative objective can improve discrimination among the alternatives encountered during training---geometrically, by enlarging Target--Sibling margins---but in our setting it spreads the compact neighborhoods that generic retrieval uses, and it does not yield a robust answerability capability on a shifted table distribution. Likewise, a general query--table relevance objective does not substitute for target-versus-sibling answerability supervision, and a simple data mixture does not automatically reconcile the two.
These findings refine the scope of our main claim within the setups above: \emph{among the generic semantic retrievers and contrastive recipes we evaluate, an answerability bottleneck persists, and straightforward answerability-oriented fine-tuning---including a simple TCR-Bench+TARGET mixture---does not provide an easy resolution, because answerability discrimination and broad semantic relevance are not automatically aligned}. Developing training objectives that jointly support both properties remains an open direction.

\section{Statistical Significance Analysis}
\label{sec:appendix_significance}

Unless otherwise stated, all reported 95\% confidence intervals for DS@1 and $p$-values are estimated via bootstrap resampling over queries ($N=209$, 1{,}000 samples, percentile method). To evaluate the statistical reliability of the observed improvements, we conduct paired bootstrap significance tests comparing answerability-aware reranking variants against the embedding-only Qwen3-E8B retrieval baseline.

\paragraph{Results}

Table~\ref{tab:significance_results} summarizes the statistical estimates for representative retrieval and reranking models. Table~\ref{tab:significance_rerank} reports paired bootstrap comparisons between AAR variants and the Qwen3-E8B baseline; note that AAR-CE is based on Qwen3-R8B, whereas the other AAR variants are based on Qwen3-30B.

\begin{table*}[htbp]
\centering
\small
\caption{Bootstrap confidence intervals and significance tests for representative retrieval methods.}
\label{tab:significance_results}
\resizebox{\textwidth}{!}{
\begin{tabular}{l|l|ccccc}
\toprule
\textbf{Model} & \textbf{Type} & \textbf{R@1} & \textbf{DS@1} & \textbf{DS@1 (95\% CI)} & \textbf{p-value vs.\ Random} \\
\midrule
tapas\_nq\_retriever\_large & DTR & 0.005 & 1.000 & [0.000, 1.000] & 0.350 \\
tapas\_nq\_hn\_retriever\_large & DTR & 0.010 & 0.250 & [0.000, 0.600] & 0.628 \\
SauerkrautLM-Multi-ModernColBERT & ColBERT-style & 0.173 & 0.225 & [0.164, 0.291] & 0.985 \\
reason-colBERT-150M-GTE-ModernColBERT & ColBERT-style & 0.086 & 0.198 & [0.116, 0.286] & 0.982 \\
Qwen3-E4B + BM25 hybrid & Hybrid & 0.215 & 0.312 & [0.239, 0.389] & 0.369 \\
\midrule
Qwen3-Embedding-8B & Dense & 0.182 & 0.271 & [0.200, 0.346] & 0.748 \\
Qwen3-Embedding-4B & Dense & 0.172 & 0.252 & [0.185, 0.322] & 0.904 \\
stella\_en\_1.5B\_v5 & Dense & 0.096 & 0.182 & [0.117, 0.257] & 0.999 \\
\midrule
AAR-CE(T10) & Reranker & 0.574 & 0.750 & [0.682, 0.817] & $< 0.001$ \\
AAR-LLM-PW(T10) & Reranker & 0.536 & 0.704 & [0.636, 0.774] & $< 0.001$ \\
AAR-LLM-LW(T10) & Reranker & 0.450 & 0.584 & [0.503, 0.660] & $< 0.001$ \\
\bottomrule
\end{tabular}
}
\end{table*}

\begin{table*}[t]
\centering
\small
\caption{Paired bootstrap significance tests comparing AAR variants with the embedding-only Qwen3-E8B retrieval baseline. Confidence intervals are computed from 1{,}000 bootstrap samples.}
\label{tab:significance_rerank}
\begin{tabular}{lccccc}
\toprule
\textbf{Comparison} & $\Delta$R@1 & \textbf{95\% CI} & $\Delta$DS@1 & \textbf{95\% CI} & \textbf{$p$-value} \\
\midrule
AAR-CE(T10) $-$ Qwen3-E8B 
& +0.393 & [0.316, 0.469] 
& +0.480 & [0.377, 0.582] 
& $< 0.001$ \\

AAR-LLM-PW(T10) $-$ Qwen3-E8B 
& +0.353 & [0.273, 0.435] 
& +0.434 & [0.336, 0.539] 
& $< 0.001$ \\

AAR-LLM-LW(T10) $-$ Qwen3-E8B
& +0.268 & [0.206, 0.330]
& +0.313 & [0.232, 0.392]
& $< 0.001$ \\
\bottomrule
\end{tabular}
\end{table*}

\paragraph{Analysis}

Table~\ref{tab:significance_results} tests DS@1 against a random-selection baseline. For the dense retrievers and the alternative first-stage methods listed above, the 95\% CIs include or lie below that baseline, and none of the $p$-values indicate a significant improvement over chance. For DTR, the DS@1 estimates are additionally unstable because group recall is near zero, which is reflected in the wide intervals. The three AAR variants are the exception: all have $p<0.001$ against random selection, and their DS@1 CIs lie well above the baseline.

Table~\ref{tab:significance_rerank} compares each AAR variant with Qwen3-E8B on the same queries. The paired differences in R@1 and DS@1 are positive, and all three tests yield $p<0.001$. As noted above, AAR-CE uses Qwen3-R8B (same scale and initialization as Qwen3-E8B), whereas AAR-LLM-PW and AAR-LLM-LW use Qwen3-30B. These tests corroborate the improvements already reported.

\section{Summary of Experimental Observations}

Table~\ref{tab:findings} summarizes the major experimental observations regarding answerability-aware table retrieval and the Semantic-Answerability Gap, as observed for the models evaluated in this study.

\begin{table}[t]
\centering
\small
\caption{Summary of experimental observations on answerability-aware table retrieval and the Semantic-Answerability Gap.}
\label{tab:findings}
\renewcommand{\arraystretch}{1.2}
\begin{tabular}{p{5.5cm}|p{6.5cm}|c}
\toprule
\textbf{Observations} & \textbf{Implications for Answerability Retrieval} & \textbf{\S} \\
\midrule

The strongest dense retriever reaches only $18.2\%$ Top-1 Target Table retrieval despite consistently reaching the correct Sibling Table Group
&
The evaluated dense retrievers capture coarse semantic neighborhoods well, but struggle with the fine-grained discrimination needed to identify the uniquely answerable source
&
\ref{sec:main_results_R}
\\

\midrule

QA performance remains close to the oracle when the Target Table is ranked first, while adding more retrieved tables provides limited gains and can introduce interference
&
Retrieving the answerable source at high ranks is critical for downstream QA; additional semantically related but non-answerable sources can interfere with reasoning
&
\ref{sec:main_results_QA}
\\
\midrule

Retrieval remains broadly stable across table serializations and semantically equivalent query formulations
&
The observed Semantic-Answerability Gap is not primarily an artifact of surface formatting or query wording in our evaluation
&
\ref{sec:surface_variations}
\\
\midrule

Complete row representations generally outperform representations retaining only a sufficient subset of attributes
&
The evaluated dense retrievers are more responsive to the amount of semantic content than to whether the representation contains only the information logically sufficient for identification
&
\ref{sec:retrieval_behavior}
\\
\midrule

Row-level retrieval substantially outperforms retrieval over multi-row chunks
&
The evaluated dense embeddings can capture local row-level content, but aggregating multiple rows can dilute the information most relevant to the query
&
\ref{sec:retrieval_behavior}
\\
\midrule

Ori and RID both outperform RID+Shuf, even though RID+Shuf retains explicit row indices
&
The evaluated dense embeddings remain relatively insensitive to fine-grained row--column bindings that can determine answerability
&
\ref{sec:retrieval_behavior}
\\
\midrule

Target$+$Sibling groups form compact, low-dimensional neighborhoods, while the global embedding space remains broadly dispersed
&
For the models we evaluate, the discrimination failure is localized within semantically coherent neighborhoods rather than caused by global embedding collapse
&
\ref{sec:diagnostic_analysis}
\\
\midrule

Target--Sibling margins are consistently negative and correlate more strongly with $R@1$ than with $GR@1$
&
Remaining retrieval error concentrates on query-to-table discrimination after the correct semantic neighborhood has been reached
&
\ref{sec:diagnostic_analysis}
\\
\midrule

Query-conditioned answerability-aware reranking substantially improves $R@1$, from $0.182$ to $0.450$--$0.574$
&
Explicit answerability assessment can distinguish the Target from semantically similar Sibling Tables more effectively than single-vector retrieval in our setting
&
\ref{sec:answerability_modeling}
\\
\midrule

Answerability-oriented fine-tuning substantially improves in-distribution retrieval but degrades across independent datasets and can hurt broader semantic retrieval
&
Answerability can be learned by the embedding model we fine-tune, but jointly preserving answerability sensitivity and broad semantic retrieval remains challenging under the recipes we explore
&
\ref{sec:answerability_modeling}
\\

\bottomrule
\end{tabular}
\end{table}

\end{document}